\documentclass[letterpaper,journal]{IEEEtran}

\usepackage{amsmath, amssymb, amsfonts}
\usepackage{graphicx}
\usepackage[nocompress]{cite}
\usepackage{booktabs}
\usepackage{multirow} 
\newtheorem{assumption}{Assumption}
\usepackage[hidelinks]{hyperref}

\begin{document}
\bstctlcite{IEEErefcontrol}

\title{\fontsize{24}{28}\selectfont Energy-Based Fault Detection and Isolation\\ Framework for Flat Port-Hamiltonian\\ Mechanical Systems}

\author{Johnny~Garcia, 
        Nicolas~Salguero, 
        Naveed~Iqbal, 
        Brahim~Brahmi,
        and~Ramy~Rashad%
\thanks{Manuscript dated 10 October 2026. This work was supported by King Fahd University of Petroleum and Minerals (KFUPM). \textit{(Corresponding author: Johnny Garcia.)}}%
\thanks{Johnny Garcia and Nicolas Salguero are with the Electrical Engineering Department, King Fahd University of Petroleum and Minerals, Dhahran 31261, Saudi Arabia (e-mail: g202427540@kfupm.edu.sa; g202426740@kfupm.edu.sa).}%
\thanks{Naveed Iqbal is with the Electrical Engineering Department and the Interdisciplinary Research Center for Communication Systems and Sensing, King Fahd University of Petroleum and Minerals, Dhahran 31261, Saudi Arabia (e-mail: naveediqbal@kfupm.edu.sa).}%
\thanks{Brahim Brahmi is with the Electrical Engineering Department and the Interdisciplinary Research Center for Intelligent Manufacturing and Robotics, King Fahd University of Petroleum and Minerals, Dhahran 31261, Saudi Arabia (e-mail: brahim.brahmi@kfupm.edu.sa).}%
\thanks{Ramy Rashad is with the Control and Instrumentation Engineering Department and the Interdisciplinary Research Center for Smart Mobility and Logistics, King Fahd University of Petroleum and Minerals, Dhahran 31261, Saudi Arabia (e-mail: ramy.rashad@kfupm.edu.sa).}%
}
\maketitle

\begin{abstract}
Fault diagnosis during regulated motion requires informative signals even when feedback attenuates tracking deviations. This paper develops an energy-based diagnostic framework that constructs mechanical port-power residuals without a dedicated dynamic observer. Spectral Hamiltonian Algebraic Power Evaluation (SHAPE) combines port-Hamiltonian variables and differential flatness to generate nominal power references from planned motion for fully actuated mechanical systems with collocated actuation. Subtracting these references from commanded effort multiplied by measured flow yields residuals whose final evaluation requires neither recursive observer updates nor measured-state dynamics calculations. Continuous wavelet transform (CWT) scalograms and a shared-encoder convolutional neural network assess their ability to distinguish nominal operation, actuator loss of effectiveness, anomalous friction, and vibration from complete trajectories. Evaluation uses a simulated six-degree-of-freedom Kinova manipulator under model predictive control and fixed plant-model mismatch, with single faults at a specified joint. Across five training seeds and \textbf{3000} held-out test runs, SHAPE--CWT achieves a mean accuracy of \textbf{84.69 percent}, versus \textbf{81.67 percent} for torque residuals and \textbf{97.17 percent} for the generalized momentum observer under common CWT processing. The respective sample standard deviations are \textbf{0.70}, \textbf{0.70}, and \textbf{0.37} percentage points. These results demonstrate useful fault-type discrimination from algebraic power residuals, while severity--excitation profiles and offline velocity perturbations reveal limits to their diagnostic capability.
\end{abstract}


\section{Introduction}

\IEEEPARstart{F}{ault} detection and isolation (FDI) supports the reliable operation of mechanical systems by identifying faults during commanded motion. This task is challenging under feedback control; the controller can compensate for degradation and disturbances while maintaining tracking performance, masking fault signatures in the tracking error~\cite{sconyers2013masked}. Small tracking deviations can therefore coexist with abnormal operation, creating a need for diagnostic signals that remain informative during regulated motion.

Data-driven methods learn fault signatures from measured signals and their representations. Convolutional neural networks (CNNs), combined with time--frequency representations such as continuous wavelet transform (CWT) scalograms, are widely used to extract diagnostic patterns in machinery~\cite{Neupane2025DataDriven,salguero2026}. Their performance depends on the information available in the selected inputs. Physics-guided diagnosis incorporates model knowledge into the learning pipeline~\cite{Wu2024PhysicsInformed}, motivating the construction of diagnostic inputs that express discrepancies from physically defined nominal behavior.

Model-based residuals provide such discrepancies. Generalized-momentum observers support actuator FDI and sensorless collision detection without acceleration measurements~\cite{deluca2003actuator, deluca2005sensorless}, and residual-driven observers can reconstruct actuator fault profiles~\cite{deluca2005actuator_identification}. For sufficiently accurate trajectory tracking, another approach compares measured joint torque with the nominal torque calculated along the desired motion~\cite{haddadin2008collision}.

An energetic perspective relates diagnostic signals to power exchange, storage, and dissipation. The port-Hamiltonian (pH) framework makes these relationships explicit through interconnected physical subsystems and effort--flow ports~\cite{vanderschaft2014, rashad2020twenty}. Energy-aware control illustrates the value of this perspective: energy tanks can regulate energy exchange to support stable and safe interaction~\cite{rashad2022energyaware}. For diagnosis, unknown-input Hamiltonian observers generate residuals and estimate faults in linear pH systems under observer-existence and disturbance-decoupling conditions~\cite{atitallah2015}. Mechanical energy observers also detect external interactions through a dynamically updated energy discrepancy~\cite{haddadin2017robot}. These developments motivate exploiting energetic structure to construct diagnostic signals, while exploring how much of this task can be achieved without a dedicated dynamic observer.

Algebraic diagnosis and power monitoring provide complementary precedents. Differential flatness supplies analytical redundancy through discrepancies between measurements and reconstructed states or inputs~\cite{saied2019flatness}. Polynomial-approximation operators permit algebraic reconstruction of residuals or additive faults, with filtering and computable error bounds~\cite{lomakin2022reliable}. Power-consumption diagnosis compares healthy electrical-consumption profiles with measurements and combines model fitting with neural-network classification~\cite{sabry2020power}.

The objective pursued here is to use mechanical port power for fault diagnosis without synthesizing a dedicated dynamic observer. For systems with planned motion, differential flatness parameterizes nominal states and inputs through reference outputs and their finite derivatives~\cite{fliess1995flatness, siraramirez2004}, offering a basis for algebraic nominal power references. Establishing which fault types these power discrepancies can distinguish under feedback, model mismatch, and measurement errors is essential to assessing their diagnostic value.

To address this objective, this paper introduces the Spectral Hamiltonian Algebraic Power Evaluation (SHAPE) framework for fully actuated, differentially flat pH mechanical systems with collocated actuation ports. SHAPE combines pH port variables and flatness-based nominal references to construct a power residual at each actuation port. Subtracting nominal port power from commanded effort multiplied by measured port flow yields physically interpretable diagnostic inputs. With nominal references available, residual evaluation requires neither recursive observer updates nor dynamics evaluation at the measured state.

The diagnostic realization maps synchronized port residuals into CWT scalograms and processes them with a shared-encoder CNN. The evaluation classifies complete trajectories of a simulated six-degree-of-freedom (6-DoF) manipulator under flatness-based model predictive control (MPC), with single faults injected at a specified joint.

The main contributions are as follows:
\begin{enumerate}
    \item \textbf{Algebraic port-wise power residuals:} A nominal reference generator based on pH port variables and differential flatness, together with a commanded-power residual and a decomposition organizing fault, dynamic-mismatch, and measurement contributions under the stated assumptions.
    \item \textbf{Physics-guided diagnostic realization:} A CWT and shared-encoder CNN pipeline using synchronized port-power residuals to distinguish nominal operation, actuator loss of effectiveness (LoE), anomalous friction, and oscillatory disturbances.
    \item \textbf{Comparative diagnostic characterization:} An evaluation of four residual formulations under common CWT processing and three complete SHAPE representation pipelines, distinguishing correct fault-type identification, fault-type confusion, and missed detection. Joint severity--excitation profiles and complementary vibration experiments characterize where diagnostic coverage is retained or limited, while offline sensor perturbations reveal changes in failure mode. These empirical profiles are interpreted alongside residual-construction requirements.
\end{enumerate}

Section~\ref{sec:preliminaries} presents the mathematical preliminaries, and Section~\ref{sec:shape_methodology} develops SHAPE. Sections~\ref{sec:simulation_setup} and~\ref{sec:results} describe the simulation protocols and diagnostic results, respectively. Section~\ref{sec:conclusion} concludes the paper.

\section{Mathematical Preliminaries}
\label{sec:preliminaries}

\subsection{Port-Hamiltonian Framework and Power Balance}
\label{sec:ph_framework}

The port-Hamiltonian framework models complex nonlinear physical structures through geometric and energetic principles. Power-preserving interconnections governed by a Dirac structure route energy among internal storage elements and interaction ports, whereas resistive elements account for internal power dissipation~\cite{vanderschaft2014, rashad2020twenty}. Unlike coordinate-dependent state-space formulations where energetic conservation is implicit, the pH framework provides an explicit structural separation between conservative power routing and dissipation.

Geometrically, the state space of an explicit input-state-output pH system is defined on a smooth manifold $\mathcal{X}$ of finite dimension $n_x$, equipped with a smooth Hamiltonian function bounded below, $H: \mathcal{X} \rightarrow \mathbb{R}_{\ge 0}$ representing the total stored energy. Physical state flows $\dot{x}$ belong to the tangent space $T_x\mathcal{X}$, whereas dual efforts belong to the cotangent space $T_x^*\mathcal{X}$, represented in local coordinates by the co-energy vector $(\partial H/\partial x)(x) \in \mathbb{R}^{n_x}$. In the absence of direct input-to-output feedthrough, the continuous-time explicit input-state-output pH formulation maps co-energy variables and external inputs to state derivatives and conjugate outputs as
\begin{equation}
\label{eq:ph_sys}
\begin{aligned}
\dot{x} &= [J(x) - R(x)] \frac{\partial H}{\partial x}(x) + g(x)u, \\
y &= g^\top(x) \frac{\partial H}{\partial x}(x).
\end{aligned}
\end{equation}
Here, $x \in \mathcal{X}$ denotes the $n_x$-dimensional state vector, $u \in \mathbb{R}^m$ represents the input port variables, and $y \in \mathbb{R}^m$ denotes the conjugate output variables across $m$ interaction ports.

The internal matrices map co-energy variables to state flows within the system topology: the skew-symmetric interconnection matrix $J(x) = -J^\top(x) \in \mathbb{R}^{n_x \times n_x}$ defines the conservative power routing without net storage or generation, the symmetric positive semi-definite matrix $R(x) = R^\top(x) \ge 0 \in \mathbb{R}^{n_x \times n_x}$ characterizes internal energy dissipation, and the input matrix $g(x) \in \mathbb{R}^{n_x \times m}$ establishes the port interaction topology coupling the system with its environment.

Evaluating the total time derivative of the Hamiltonian along system trajectories reveals the energetic behavior. By the chain rule, $\dot{H} = \left((\partial H/\partial x)(x)\right)^\top \dot{x}$. Substituting~\eqref{eq:ph_sys}, the conservative term strictly vanishes due to skew-symmetry, $\left((\partial H/\partial x)(x)\right)^\top J(x) (\partial H/\partial x)(x) \equiv 0$, yielding the fundamental power balance equation
\begin{equation}
\label{eq:power_balance}
\dot{H} = y^\top u - \left(\frac{\partial H}{\partial x}(x)\right)^\top R(x) \frac{\partial H}{\partial x}(x) \le y^\top u.
\end{equation}

Inequality~\eqref{eq:power_balance} defines system passivity with respect to the supply rate $s(u, y) = y^\top u$ and storage function $H(x) \ge 0$, establishing that the rate of change of stored energy is bounded by the net power supplied through the interaction ports.

\subsection{Port-Hamiltonian Modeling of Mechanical Systems}
\label{sec:ph_mechanical}

For an $n_q$-degree-of-freedom mechanical system, the Hamiltonian function $H: T^*\mathcal{Q} \rightarrow \mathbb{R}_{\ge 0}$ describes the total stored physical energy through
\begin{equation}
\label{eq:mech_hamiltonian}
H(q, p) = \frac{1}{2}p^\top M^{-1}(q)p + V(q).
\end{equation}
Here, $\mathcal{Q}$ denotes an $n_q$-dimensional smooth configuration manifold, $q \in \mathbb{R}^{n_q}$ represents the vector of generalized position coordinates in a local chart, $M(q) = M^\top(q) > 0$ is the generalized inertia matrix, and $V(q)$ denotes the potential energy field (calibrated such that $\min_{q} V(q) = 0$). The canonical momentum vector $p \in T_q^*\mathcal{Q}$ is defined on the cotangent fiber via the Legendre transformation as $p = M(q)\dot{q}$.

In local phase-space coordinates $x = [q^\top, p^\top]^\top \in \mathbb{R}^{n_x}$ with $n_x = 2n_q$, the co-energy vector partitions as
\begin{equation}
\label{eq:coenergy_vector}
\frac{\partial H}{\partial x}(x) = \begin{bmatrix} \frac{\partial H}{\partial q}(q,p) \\ \frac{\partial H}{\partial p}(q,p) \end{bmatrix} = \begin{bmatrix} \frac{\partial V}{\partial q}(q) + \frac{1}{2} \frac{\partial}{\partial q}\left(p^\top M^{-1}(q)p\right) \\ M^{-1}(q)p \end{bmatrix}.
\end{equation}
This identifies the generalized velocity vector as $\dot{q} = (\partial H/\partial p) = M^{-1}(q)p$. To couple this mechanical state space with resistive dissipation and external interaction ports, the structure matrices from Section~\ref{sec:ph_framework} instantiate into canonical mechanical blocks:
\begin{equation}
\label{eq:ph_matrices_block}
\begin{aligned}
J_0 &= \begin{bmatrix} 0_{n_q} & I_{n_q} \\ -I_{n_q} & 0_{n_q} \end{bmatrix}, \quad g(x) = \begin{bmatrix} 0_{n_q \times m} \\ B(q) \end{bmatrix}, \\[6pt]
R(x) &= \begin{bmatrix} 0_{n_q} & 0_{n_q} \\ 0_{n_q} & D(x) \end{bmatrix}.
\end{aligned}
\end{equation}
Here, $J_0 = -J_0^\top \in \mathbb{R}^{n_x \times n_x}$, $R(x) = R^\top(x) \ge 0 \in \mathbb{R}^{n_x \times n_x}$ with $D(x) \in \mathbb{R}^{n_q \times n_q}$ ($D(x) = D^\top(x) \succeq 0$) modeling state-dependent internal dissipation (such as viscous and friction phenomena), and $B(q) \in \mathbb{R}^{n_q \times m}$ is the input coupling matrix for $m$ independent actuation ports with $m \le n_q$ (non-redundant actuation).

Substituting~\eqref{eq:ph_matrices_block} and~\eqref{eq:coenergy_vector} into~\eqref{eq:ph_sys} yields the mechanical equations of motion in phase space under impedance causality:
\begin{equation}
\label{eq:mech_ph_dynamics}
\begin{aligned}
\dot{q} &= M^{-1}(q)p, \\
\dot{p} &= -\frac{\partial H}{\partial q}(q,p) - D(x)M^{-1}(q)p + B(q)u.
\end{aligned}
\end{equation}
Here, $u(t) = \tau(t) \in \mathbb{R}^m$ represents the generalized control effort (actuator torques or forces), and the conjugate port flow vector satisfies $y(t) = g^\top(x)(\partial H/\partial x)(x) = B^\top(q)\dot{q} \in \mathbb{R}^m$.

\subsection{Differential Flatness}
\label{sec:flatness}

Differential flatness establishes that the state vector $x \in \mathbb{R}^{n_x}$ and control input vector $u \in \mathbb{R}^m$ of a nonlinear dynamical system can be completely parameterized by a flat output vector $z \in \mathbb{R}^m$ and a finite number of its time derivatives. Formally, a nonlinear system is differentially flat if there exist smooth mappings $\Phi$ and $\Psi$ such that
\begin{equation}
\label{eq:flat_state}
x = \Phi(z, \dot{z}, \ddot{z}, \dots, z^{(r_x)}).
\end{equation}
\begin{equation}
\label{eq:flat_input}
u = \Psi(z, \dot{z}, \ddot{z}, \dots, z^{(r_u)}).
\end{equation}
Here, $r_x, r_u \in \mathbb{Z}_{\ge 0}$ denote the maximum finite derivation orders associated with the state and input manifolds, respectively~\cite{fliess1995flatness, siraramirez2004}. These orders depend on the chosen flat parameterization.

\section{The SHAPE Methodology}
\label{sec:shape_methodology}

The SHAPE framework constructs diagnostic inputs for differentially flat pH systems from planned nominal power and measured port flow. It evaluates the discrepancy between nominal port power and commanded effort multiplied by measured flow, without recursive observer updates in the final residual mapping. Departures from the nominal model can appear in this residual; their diagnostic separability is evaluated experimentally below.

\begin{figure}[!t]
    \centering
    \includegraphics[width=\linewidth]{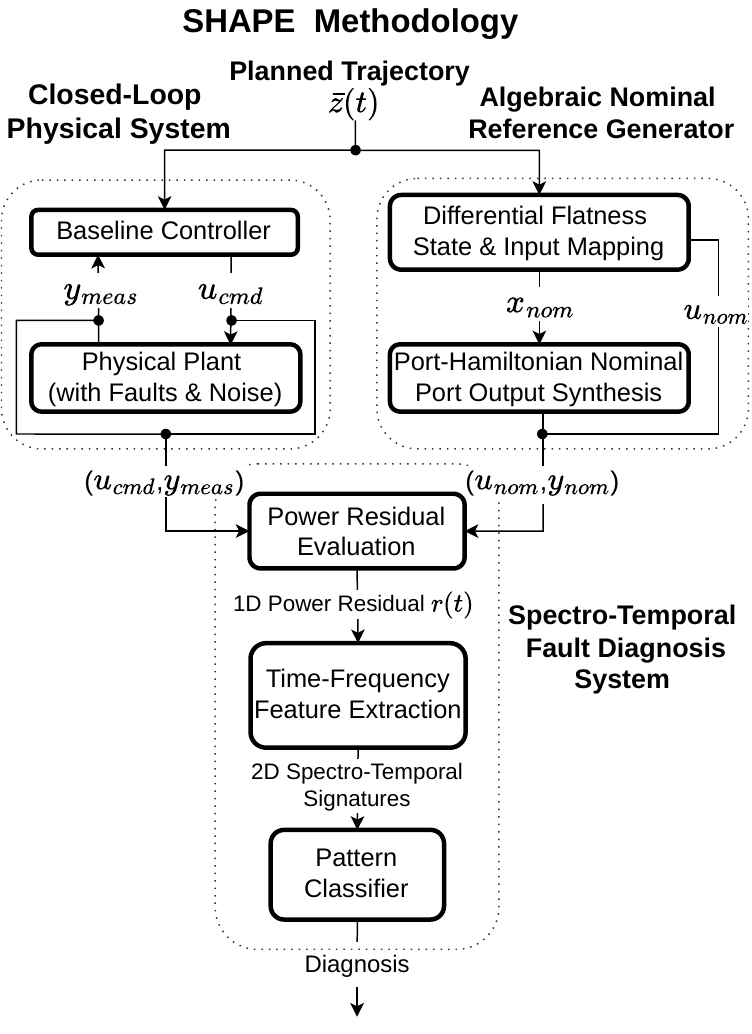}
    \caption{Conceptual architecture and energetic signal flow of the general SHAPE framework. \textbf{(Top-Right)} Algebraic Nominal Reference Generator exploits differential flatness to compute an algebraic nominal port-power baseline directly from planned trajectories. \textbf{(Top-Left)} Closed-Loop Physical System evaluates the commanded-power proxy under feedback control, plant mismatch, disturbances, and physical anomalies. \textbf{(Bottom)} Spectro-Temporal Fault Diagnosis System evaluates the port-power residual vector and maps the resulting 1-D non-stationary residuals into 2-D spectro-temporal signatures for fault diagnosis.}
    \label{fig:shape_framework}
\end{figure}

To diagnose these anomalies, the conceptual architecture of the SHAPE framework is structured into three interacting domains, as illustrated in Fig.~\ref{fig:shape_framework}: the algebraic nominal reference generator, the closed-loop physical system, and the spectro-temporal fault diagnosis system.

\subsection{Algebraic Nominal Reference Generator} \label{sec:algebraic_generator}

Exploiting the differential flatness parameterizations \eqref{eq:flat_state} and \eqref{eq:flat_input}, the nominal state vector $x_{\mathrm{nom}}(t)$ and nominal control input vector $u_{\mathrm{nom}}(t)$ are evaluated algebraically along a planned flat reference trajectory $z_{\mathrm{nom}}(t) \in C^{\max(r_x, r_u)}$ as
\begin{equation} \label{eq:nom_states}
\begin{aligned}
x_{\mathrm{nom}}(t) &= \Phi\big(z_{\mathrm{nom}}, \dot{z}_{\mathrm{nom}}, \dots, z_{\mathrm{nom}}^{(r_x)}\big), \\
u_{\mathrm{nom}}(t) &= \Psi\big(z_{\mathrm{nom}}, \dot{z}_{\mathrm{nom}}, \dots, z_{\mathrm{nom}}^{(r_u)}\big).
\end{aligned}
\end{equation}
Substituting $x_{\mathrm{nom}}(t)$ into the conjugate port-flow map
of the nominal pH model, with Hamiltonian $H_{\mathrm{nom}}$, yields
the synthesized nominal flow vector:
\begin{equation} \label{eq:nom_output}
y_{\mathrm{nom}}(t)
=
g^\top(x_{\mathrm{nom}}(t))
\frac{\partial H_{\mathrm{nom}}}{\partial x}
(x_{\mathrm{nom}}(t)).
\end{equation}

To evaluate the external power exchange on a channel-by-channel basis, the formulation adopts the following canonical setting:

\begin{assumption}[Collocated and Fully Actuated Port Coordinates]
\label{assump:collocated_ports}
The mechanical pH system with canonical state
$x = [q^\top, p^\top]^\top \in \mathbb{R}^{2n_q}$ is fully actuated
in configuration ($m = n_q$), with an input matrix
$g(q) = [0_{n_q \times n_q}^\top, B^\top(q)]^\top$ satisfying
$B(q) = I_{n_q}$. Consequently, in the selected
generalized-coordinate basis, the conjugate port variables coincide
with the generalized velocity flows $y(t) = \dot{q}(t)$ and
generalized effort inputs $u(t) = \tau(t)$.
\end{assumption}

The total external power supply
$\mathcal{P}_{\mathrm{total}}(t) = y^\top(t)u(t)$ is invariant under
power-preserving coordinate transformations, whereas its individual
scalar contributions depend on the selected port-coordinate
basis~\cite{vanderschaft2014}. Under Assumption~\ref{assump:collocated_ports}, the chosen collocated basis admits the additive channel-wise representation $\mathcal{P}_{\mathrm{total}}(t) = \sum_{i=1}^{n_q} y_i(t)u_i(t)$. Accordingly, the algebraic nominal port-power baseline for the $i$-th actuation port is defined as
\begin{equation}
\label{eq:flat_power_balance}
\mathcal{P}_{\mathrm{nom},i}(t)
=
y_{\mathrm{nom},i}(t)u_{\mathrm{nom},i}(t),
\qquad
i \in \{1,\dots,n_q\}.
\end{equation}

\subsection{Closed-Loop Physical System}
\label{sec:physical_system}

To track the planned flat reference trajectory, the physical plant
operates in closed loop under an active feedback controller that
generates the commanded input effort vector
$u_{\mathrm{cmd}}(t) \in \mathbb{R}^{n_q}$ to compensate for tracking
deviations under anomalous conditions. Unlike the idealized algebraic
generator, the physical system may exhibit parameter mismatch,
disturbances, and physical anomalies, while its measured port variables
may be affected by sensor noise.

The state evolution of the perturbed physical system is modeled in
pH form as
\begin{subequations}
\label{eq:faulty_ph}
\begin{equation}
\label{eq:faulty_ph_state}
\dot{x}
=
[J_0-R_{\mathrm{flt}}(x)]
\frac{\partial H}{\partial x}(x)
+g\bigl(\Pi u_{\mathrm{cmd}}(t)+d_\tau(x,t)\bigr).
\end{equation}
Its measured conjugate port flow is
\begin{equation}
\label{eq:faulty_ph_output}
y_{\mathrm{meas}}(t)
=g^\top\frac{\partial H}{\partial x}(x)+\nu(t).
\end{equation}
\end{subequations}
Here,
$g=[0_{n_q\times n_q}^{\top},I_{n_q}^{\top}]^{\top}
\in\mathbb{R}^{2n_q\times n_q}$ is the canonical input distribution
matrix. The fault-affected resistive structure is
$R_{\mathrm{flt}}(x)
=\operatorname{diag}(0_{n_q\times n_q},D(x))\succeq0$, where
$D(x)=D_{\mathrm{nom}}(x)+\Delta D_{\mathrm{mdl}}(x)
+\Delta D_{\mathrm{flt}}(x)$, with
$D_{\mathrm{nom}}(x)+\Delta D_{\mathrm{mdl}}(x)\succeq0$
representing the healthy physical dissipation,
$\Delta D_{\mathrm{mdl}}(x)$ accounting for the persistent
plant-to-reference model mismatch, and
$\Delta D_{\mathrm{flt}}(x)\succeq0$ representing additional
fault-induced dissipation acting on the momentum coordinates. The
diagonal matrix
$\Pi=\operatorname{diag}(\alpha_1,\dots,\alpha_{n_q})$, with
$\alpha_i\in(0,1]$, models actuator LoE while
preserving full input rank. Finally,
$d_\tau(x,t)\in\mathbb{R}^{n_q}$ is an additive exogenous effort
disturbance acting on the momentum dynamics, and
$\nu(t)\in\mathbb{R}^{n_q}$ represents zero-mean measurement noise.

To formalize the structural integrity of the plant under this
degradation class, the framework establishes the following condition:

\begin{assumption}[Topological Invariance and Kinematic Integrity]
\label{assump:topological_invariance}
The physical anomaly set
$\{\Pi,\Delta D_{\mathrm{flt}},d_\tau\}$ alters actuator effectiveness,
internal dissipation, or exogenous effort, but preserves the continuous
configuration manifold $\mathcal{Q}$ of dimension $n_q$ and the
canonical mechanical pH structure established in
Section~\ref{sec:ph_mechanical} under
Assumption~\ref{assump:collocated_ports}.
\end{assumption}

Because the delivered actuator-port power
$\alpha_i y_i(t)u_{\mathrm{cmd},i}(t)$ is not directly measured, the
diagnostic architecture evaluates the \emph{commanded port-power proxy}
$\mathcal{P}_{\mathrm{cmd},i}(t)\in\mathbb{R}$:
\begin{equation}
\label{eq:meas_power}
\mathcal{P}_{\mathrm{cmd},i}(t)
\triangleq
y_{\mathrm{meas},i}(t)u_{\mathrm{cmd},i}(t),
\qquad
i\in\{1,\dots,n_q\}.
\end{equation}
This commanded proxy is deliberate: unlike a delivered-effort
measurement, which already reflects actuator degradation, evaluating
the commanded effort retains the command-to-delivery power discrepancy
$(1-\alpha_i)y_i(t)u_{\mathrm{cmd},i}(t)$ at the diagnostic port.

\subsection{Spectro-Temporal Fault Diagnosis System} \label{sec:diagnosis_system}

\subsubsection{Power Residual Evaluation} \label{sec:residual_evaluation}
For each actuation port channel $i \in \{1, \dots, n_q\}$, the diagnostic architecture evaluates a scalar energetic residual $r_{\mathrm{SHAPE},i}(t) \in \mathbb{R}$, defined as the discrepancy between the commanded port power proxy \eqref{eq:meas_power} and the nominal algebraic baseline \eqref{eq:flat_power_balance}:
\begin{equation} \label{eq:residual_def}
\begin{aligned}
r_{\mathrm{SHAPE},i}(t) &\triangleq \mathcal{P}_{\mathrm{cmd}, i}(t) - \mathcal{P}_{\mathrm{nom}, i}(t).
\end{aligned}
\end{equation}

To relate this discrepancy to explicit physical contributions while retaining the coupled Hamiltonian dynamics, we project the perturbed system \eqref{eq:faulty_ph} onto the $i$-th coordinate momentum:
\begin{equation} \label{eq:p_dot_coord}
\dot{p}_i(t) = -\left[ \frac{\partial H}{\partial q}(q, p) \right]_i - [D(x)y]_i + \alpha_i u_{\mathrm{cmd}, i}(t) + d_{\tau, i}(x, t).
\end{equation}
Multiplying by the true generalized flow $y_i(t) = \dot{q}_i(t)$ yields the actuator-port power relation $\alpha_i y_i u_{\mathrm{cmd}, i} = \mathcal{P}_{\mathrm{mec}, i} + y_i[D(x)y]_i - y_i d_{\tau, i}$, where
\begin{equation} \label{eq:P_mech_coord}
\mathcal{P}_{\mathrm{mec}, i}(t) \triangleq y_i(t) \left( \dot{p}_i(t) + \left[ \frac{\partial H}{\partial q}(q(t), p(t)) \right]_i \right).
\end{equation}
The quantities $\mathcal{P}_{\mathrm{mec}, i}(t)$ are coordinate contributions to the global Hamiltonian rate, whose sum satisfies $\sum_{i=1}^{n_q} \mathcal{P}_{\mathrm{mec}, i}(t) \equiv \dot{H}(q(t), p(t))$, with individual values depending on the selected generalized-coordinate basis. Expanding \eqref{eq:meas_power} via
$u_{\mathrm{cmd},i}
=
\alpha_i u_{\mathrm{cmd},i}
+
(1-\alpha_i)u_{\mathrm{cmd},i}$ with
$y_{\mathrm{meas},i}=y_i+\nu_i$, and subtracting the nominal planned
baseline
$\mathcal{P}_{\mathrm{nom},i}
=
\mathcal{P}_{\mathrm{mec,nom},i}
+
y_{\mathrm{nom},i}
[D_{\mathrm{nom}}(x_{\mathrm{nom}})
y_{\mathrm{nom}}]_i$,
where $\mathcal{P}_{\mathrm{mec,nom},i}$ denotes the nominal
counterpart of \eqref{eq:P_mech_coord}, the scalar residual
\eqref{eq:residual_def} decomposes directly into five constituent
terms in watts ($\mathrm{W}$):
\begin{equation} \label{eq:residual_decomposition}
\begin{aligned}
r_{\mathrm{SHAPE},i}(t) &= (1 - \alpha_i) y_i(t) u_{\mathrm{cmd}, i}(t) + y_i(t) [\Delta D_{\mathrm{flt}}(x) y(t)]_i \\
&\quad - y_i(t) d_{\tau, i}(x, t) + e_{\mathrm{dyn}, i}(t) + \nu_i(t) u_{\mathrm{cmd}, i}(t).
\end{aligned}
\end{equation}
The coordinate dynamic mismatch is defined as
\begin{align}
\label{eq:edyn_def}
e_{\mathrm{dyn},i}(t)
&\triangleq
\mathcal{P}_{\mathrm{mec},i}(t)
-
\mathcal{P}_{\mathrm{mec,nom},i}(t)
\notag\\
&\quad
+
y_i(t)
\bigl[
\bigl(D_{\mathrm{nom}}(x)
+
\Delta D_{\mathrm{mdl}}(x)\bigr)y(t)
\bigr]_i
\notag\\
&\quad
-
y_{\mathrm{nom},i}(t)
[D_{\mathrm{nom}}(x_{\mathrm{nom}})
y_{\mathrm{nom}}(t)]_i.
\end{align}

The decomposition~\eqref{eq:residual_decomposition} provides an
algebraically exact organization of the residual into physically
interpretable contributions under the stated assumptions. Because
measured flow, commanded effort, and physical trajectory are
closed-loop variables affected by the active anomaly, the magnitudes
of these contributions remain dynamically interdependent.
The term $(1-\alpha_i)y_i u_{\mathrm{cmd},i}$ represents the direct
command-to-delivery power discrepancy associated with actuator loss
of effectiveness, $-y_i d_{\tau,i}$ represents the residual
contribution associated with exogenous effort disturbances,
$e_{\mathrm{dyn},i}$ accounts for the healthy plant-to-reference
model mismatch and closed-loop dynamic mismatch, and
$y_i[\Delta D_{\mathrm{flt}}(x)y]_i$ represents the coordinate
contribution associated with anomalous internal dissipation.
For coupled dissipation, positive semi-definiteness guarantees
$y^\top\Delta D_{\mathrm{flt}}(x)y\ge0$, although an individual
coordinate contribution $y_i[\Delta D_{\mathrm{flt}}(x)y]_i$ need
not be nonnegative. Under zero-mean measurement noise
$\nu_i\sim\mathcal{N}(0,\sigma_{\nu,i}^2)$ independent of the control
input, the feedthrough satisfies
$\mathbb{E}[\nu_i u_{\mathrm{cmd},i}\mid u_{\mathrm{cmd},i}]=0$,
with conditional variance
$\operatorname{Var}[\nu_i u_{\mathrm{cmd},i}\mid
u_{\mathrm{cmd},i}]
=
\sigma_{\nu,i}^2u_{\mathrm{cmd},i}^2$,
establishing an effort-dependent heteroscedastic noise floor.

Within each scalar temporal channel, the constituent mechanisms in \eqref{eq:residual_decomposition} are superposed. Because measured flow and commanded effort evolve under closed-loop feedback, an anomaly at one port can induce transient responses across coupled actuation coordinates. Moreover, under non-stationary operation, the spectral content of the residual may evolve throughout the observation interval. These properties motivate evaluating a joint spectro-temporal representation that retains temporal localization together with spectral organization alongside the original temporal residual.

\subsubsection{Abstract Spectro-Temporal Representation}
\label{sec:tfa_extraction}

Let $\mathcal{I}=[0,T]$ denote a finite observation interval, and let
$\Lambda\subseteq\mathbb{R}$ denote an ordered spectral-coordinate
domain. For each actuation port channel
$i\in\{1,\dots,n_q\}$, the scalar power residual is treated as a
square-integrable signal $r_{\mathrm{SHAPE},i}\in L^2(\mathcal{I})$. SHAPE constructs
a two-dimensional spectro-temporal representation through the abstract
mapping
\begin{equation}
\label{eq:general_tfa_operator}
\mathcal{T}_{\mathrm{ST}}
\colon
L^2(\mathcal{I})
\to
\mathcal{H}_{\mathrm{ST}},
\qquad
r_{\mathrm{SHAPE},i}
\mapsto
Z_i(t,\lambda).
\end{equation}
Here, $\mathcal{H}_{\mathrm{ST}}$ denotes a space of representations
indexed by $(t,\lambda)\in\mathcal{I}\times\Lambda$. The interpretation
of $\lambda$ depends on the selected realization and may correspond to
frequency, scale, or another ordered spectral parameterization.

The continuous representation is converted into a finite
two-dimensional grid through the discretization operator
\begin{equation}
\label{eq:scalogram_tensor}
\mathcal{D}
\colon
\mathcal{H}_{\mathrm{ST}}
\to
\mathbb{R}^{N_\lambda\times N_b},
\qquad
\mathbf{X}_i
\triangleq
\mathcal{D}\!\left(
\mathcal{T}_{\mathrm{ST}}(r_{\mathrm{SHAPE},i})
\right).
\end{equation}
Here, $N_\lambda,N_b\in\mathbb{N}^{+}$ denote the spectral and
temporal grid dimensions, respectively. The resulting representation is
governed by the following structural design requirements:

\begin{enumerate}
    \item \textbf{Joint Spectro-Temporal Localization:}
    $Z_i(t,\lambda)$ provides coefficients jointly indexed by time
    and spectral coordinate, such that transient residual variations
    remain temporally localized and spectrally organized throughout
    the observation interval.

    \item \textbf{Ordered Two-Dimensional Morphology:}
    The discretization operator $\mathcal{D}$ preserves the ordering
    and neighborhood relations of the temporal samples
    $t_1<\cdots<t_{N_b}$ and spectral samples
    $\lambda_1<\cdots<\lambda_{N_\lambda}$. Consequently, neighboring
    entries of $\mathbf{X}_i$ correspond to neighboring temporal
    and spectral regions.

    \item \textbf{Portwise Representation Consistency:}
    Within a given SHAPE realization, the same operators
    $\mathcal{T}_{\mathrm{ST}}$ and $\mathcal{D}$, temporally aligned
    observation grid, ordered spectral grid, and preprocessing rule
    are applied to every actuation port channel.
\end{enumerate}

Together, these requirements define the class of two-dimensional
residual morphologies used by SHAPE.

\subsubsection{Pattern Classification} \label{sec:pattern_classification}
Let
$\mathbf{X}
=
\operatorname{stack}_{i=1}^{n_q}
\left(\mathbf{X}_i\right)
\in
\mathbb{R}^{n_q\times N_\lambda\times N_b}$
denote the composite tensor aggregating the spectro-temporal grids from
all active configuration coordinates, where $N_\lambda$ and $N_b$
represent the numbers of spectral and temporal bins, respectively, and
$\mathbf{X}_i\in\mathbb{R}^{N_\lambda\times N_b}$ represents the
local grid for coordinate $i$. Under a single-fault-location
formulation, the framework defines a learned, inductive classification
operator
\begin{equation} \label{eq:classification_operator}
\mathcal{C}
\colon
\mathbb{R}^{n_q\times N_\lambda\times N_b}
\to
\mathcal{F}_{\mathrm{diag}},
\qquad
\mathbf{X}\mapsto\hat{c}.
\end{equation}
Here, $\mathcal{F}_{\mathrm{diag}}$ is the discrete set of trajectory-level health classes corresponding to the nominal regime and the physical degradation modes $\{\Pi, \Delta D_{\mathrm{flt}}, d_\tau\}$ formalized in Assumption~\ref{assump:topological_invariance}, with $\hat{c}$ denoting the predicted class. This formulation restricts the operator to multi-class fault-type categorization at a fixed, predetermined coordinate, isolating fault-type identification from the separate problem of spatial fault localization across the $n_q$ configuration coordinates.

To account for the dynamic coupling and energy redistribution induced
by closed-loop feedback across the multi-port channels, the inductive
classification operator $\mathcal{C}$ is structured according to three
architectural design requirements:
\begin{enumerate}
    \item \textbf{Homogeneous Local Feature Extraction:} To enforce uniform representational capacity across configuration coordinates, the extraction of nonlinear spectro-temporal features within each local representation $\mathbf{X}_i$ is governed by shared projection parameters across channels.
    \item \textbf{Inter-Port Dynamic Cross-Correlation:}
To distinguish fault-mechanism signatures from reactive,
controller-induced power fluctuations distributed across coupled
actuation ports, the operator captures nonlinear cross-channel
dependencies across all configuration coordinates
$i\in\{1,\dots,n_q\}$.
    \item \textbf{Class-Discriminative Fusion under Feedback Masking:} To support empirical separation of the prescribed health classes under feedback attenuation, the operator fuses the multi-port energy fluctuations into a learned discriminative representation.
\end{enumerate}

Together, these architectural design requirements complete the
methodological formulation of the classification operator
$\mathcal{C}$ by defining an empirical inductive mapping from multi-port
closed-loop power residuals to the discrete trajectory-level health
classes in $\mathcal{F}_{\mathrm{diag}}$.

\section{Simulation Setup: 6-DoF Manipulator}
\label{sec:simulation_setup}

\begin{figure}[!t]
    \centering
    \includegraphics[width=\linewidth]{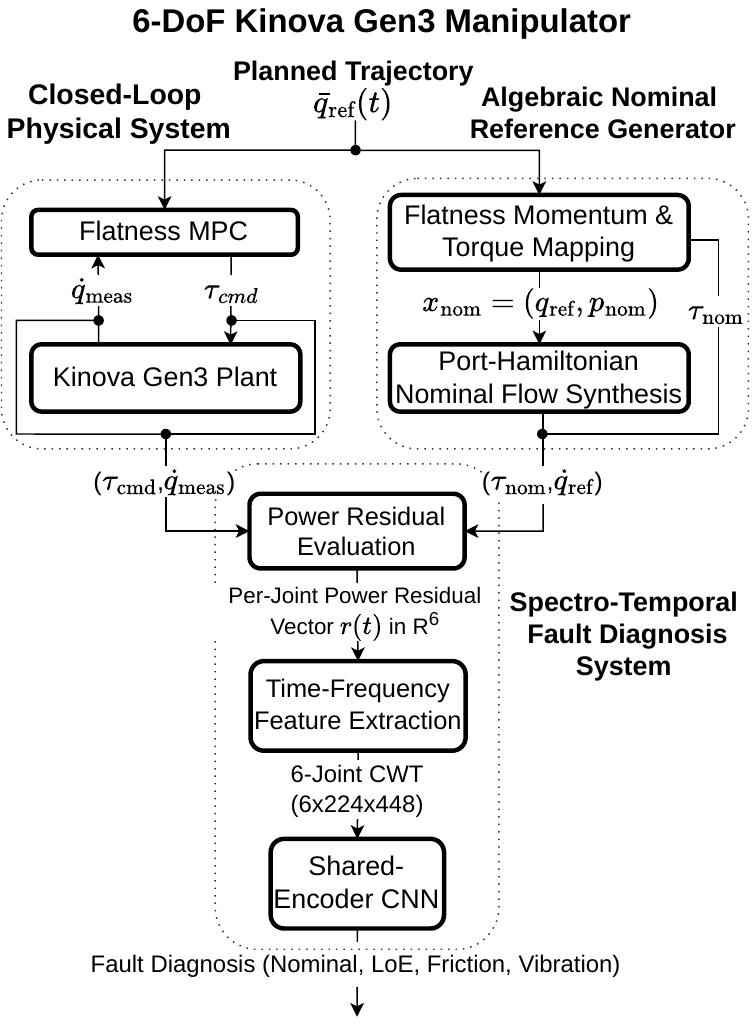}
    \caption{Simulation implementation and signal flow of the SHAPE framework on the 6-DoF Kinova Gen3 manipulator. \textbf{(Top-Right)} Algebraic Nominal Reference Generator computes the nominal joint state trajectory and baseline joint power profile via inverse dynamics along planned trajectories. \textbf{(Top-Left)} Closed-loop robotic system produces the physical telemetry and commanded joint-power proxy under Flatness-MPC, computed-torque control, model mismatch, and injected faults. \textbf{(Bottom)} Spectro-Temporal Fault Diagnosis System evaluates the per-joint power residual vector, and transforms the 1-D temporal channels via CWT into multi-joint scalogram tensors for CNN classification.}
    \label{fig:control_architecture}
\end{figure}

The SHAPE framework was instantiated and evaluated in simulation on a fully actuated 6-DoF Kinova Gen3 manipulator. Fig.~\ref{fig:control_architecture} summarizes the resulting simulation and diagnostic workflow. Section~\ref{subsec:nominal_gen} defines the reference-trajectory population and constructs the nominal joint states, torques, and port-power baselines through flatness and inverse dynamics. Section~\ref{subsec:physical_system} specifies the constrained closed-loop simulation, perturbed plant, Joint~2 anomaly models, paired physical campaign, and numerical signal acquisition. Section~\ref{subsec:spectro_temporal_dx} constructs the SHAPE and comparator residuals, their matched diagnostic representations and classifier architectures, and the training, model-selection, and statistical evaluation protocols.

\subsection{Algebraic Nominal Reference Generator and
Reference-Trajectory Population}
\label{subsec:nominal_gen}

\subsubsection{Nominal Flatness and Inverse-Dynamics Mapping}
\label{subsubsec:nominal_mapping}

For the 6-DoF Kinova Gen3 manipulator, the flat output is instantiated
as the planned joint-position trajectory
$z_{\mathrm{nom}}(t)=q_{\mathrm{ref}}(t)\in\mathbb{R}^{6}$.
The corresponding extended flat trajectory is
\begin{equation}
\label{eq:extended_z_kinova}
\bar{z}(t)
=
\left(
q_{\mathrm{ref}}(t),
\dot{q}_{\mathrm{ref}}(t),
\ddot{q}_{\mathrm{ref}}(t)
\right)
\in\mathbb{R}^{18}.
\end{equation}

Using the nominal inertia matrix
$M_{\mathrm{nom}}(q_{\mathrm{ref}})$, the nominal canonical momentum
and state trajectories are
\begin{equation}
\label{eq:nominal_momentum_kinova}
p_{\mathrm{nom}}(t)
=
M_{\mathrm{nom}}(q_{\mathrm{ref}}(t))
\dot{q}_{\mathrm{ref}}(t).
\end{equation}
\begin{equation}
\label{eq:nominal_state_kinova}
x_{\mathrm{nom}}(t)
=
\begin{bmatrix}
q_{\mathrm{ref}}^\top(t) &
p_{\mathrm{nom}}^\top(t)
\end{bmatrix}^{\top}
\in\mathbb{R}^{12}.
\end{equation}

Evaluating the nominal momentum dynamics
\eqref{eq:mech_ph_dynamics} along $\bar{z}(t)$ under the collocated
condition $B(q_{\mathrm{ref}})=I_6$ yields
$u_{\mathrm{nom}}(t)=\tau_{\mathrm{nom}}(t)$, with
\begin{equation}
\label{eq:nominal_torque_ph}
\begin{split}
\tau_{\mathrm{nom}}(t)
={}&
\dot{p}_{\mathrm{nom}}(t)
+
\frac{\partial H_{\mathrm{nom}}}{\partial q}
\bigl(x_{\mathrm{nom}}(t)\bigr)
\\
&+
D_{\mathrm{nom}}\bigl(x_{\mathrm{nom}}(t)\bigr)
\dot{q}_{\mathrm{ref}}(t).
\end{split}
\end{equation}

The nominal conjugate flow is
$y_{\mathrm{nom}}(t)=\dot{q}_{\mathrm{ref}}(t)$. Therefore, the
joint-wise nominal port-power baseline defined in
\eqref{eq:flat_power_balance} becomes
\begin{equation}
\label{eq:nominal_power_profile_joint}
\mathcal{P}_{\mathrm{nom},i}(t)
=
\dot{q}_{\mathrm{ref},i}(t)
\tau_{\mathrm{nom},i}(t),
\qquad
i\in\{1,\dots,6\}.
\end{equation}

\subsubsection{Deterministic Multi-Segment Reference Generation}
\label{subsubsec:trajectory_generation}

Each parent trajectory was defined in joint space over a horizon
$T=10~\mathrm{s}$ and partitioned into $K\in\{2,3\}$ segments of
equal duration $\Delta t=T/K$. The number of segments $K$ and the
motion mode $m\in\{\mathrm{stop},\mathrm{fly}\}$ were assigned
a priori in the experimental manifest. The temporal knots and their
associated waypoints were
\begin{equation}
\label{eq:trajectory_knots}
t_s=s\Delta t,
\quad
w_s=q_{\mathrm{ref}}(t_s)\in\mathbb{R}^{6},
\qquad
s=0,\ldots,K.
\end{equation}
Each segment was generated by quintic interpolation satisfying the
prescribed position, velocity, and acceleration conditions
\[
q_{\mathrm{ref}}(t_s)=w_s,
\qquad
\dot{q}_{\mathrm{ref}}(t_s)=v_s,
\qquad
\ddot{q}_{\mathrm{ref}}(t_s)=b_s
\]
at its boundary knots.

For both motion modes, zero acceleration was imposed at every knot,
$b_s=0$, together with zero endpoint velocities, $v_0=v_K=0$. The
internal-knot velocities were defined by
\begin{equation}
\label{eq:fly_stop_knot_conditions}
\begin{aligned}
v_s &=
\begin{cases}
0,
& m=\mathrm{stop},
\\[1mm]
\operatorname{sat}_{\beta\dot{q}_{\mathrm{lim}}}
\left(
\dfrac{w_{s+1}-w_{s-1}}{2\Delta t}
\right),
& m=\mathrm{fly},
\end{cases}
\\
&\quad s=1,\ldots,K-1;
\quad
\beta=0.65.
\end{aligned}
\end{equation}
Here,
$\operatorname{sat}_{\beta\dot{q}_{\mathrm{lim}}}(\cdot)$ denotes
componentwise saturation at
$\beta\dot{q}_{\mathrm{lim}}$. The \emph{stop} mode reaches zero
velocity at each internal knot without introducing a finite dwell
interval, whereas the \emph{fly} mode traverses the knots using bounded
centered tangents. The shared knot conditions provide $C^2$ continuity
across consecutive segments.

The waypoints were sampled sequentially. The initial waypoint was drawn
uniformly from the admissible joint set $\mathcal{Q}_{\mathrm{adm}}$,
whereas each subsequent waypoint was drawn according to
\begin{equation}
\label{eq:reachable_waypoint_sampling}
\begin{aligned}
w_0
&\sim
\mathcal{U}\!\left(\mathcal{Q}_{\mathrm{adm}}\right),
\\
w_s
&\sim
\mathcal{U}\!\left(
\mathcal{Q}_{\mathrm{adm}}
\cap
\left\{
q\in\mathbb{R}^{6}:
\left|q-w_{s-1}\right|
\preceq
\alpha\Delta t\,\dot{q}_{\mathrm{lim}}
\right\}
\right),
\\
&\quad s=1,\ldots,K;
\quad
\alpha=\frac{0.92}{1.875}.
\end{aligned}
\end{equation}
Here, $\mathcal{U}(\mathcal{A})$ denotes the uniform distribution over
the set $\mathcal{A}$ and the inequality is interpreted componentwise.
The bounds defining $\mathcal{Q}_{\mathrm{adm}}$ and
$\dot{q}_{\mathrm{lim}}$ are specified in
Section~\ref{subsubsec:trajectory_qualification}. Parent-specific
random seeds were recorded in the experimental manifest.

\subsubsection{Full-Grid Qualification and Frozen Population}
\label{subsubsec:trajectory_qualification}

Before full-grid qualification, every candidate waypoint was subjected
to the nominal static-feasibility condition
$\left|\tau_{g,\mathrm{nom}}(w_s)\right|
\preceq0.95\,\tau_{\mathrm{lim}}$ for $s=0,\ldots,K$, where
$\tau_{g,\mathrm{nom}}(w_s)\in\mathbb{R}^{6}$ is the nominal gravity
torque and
$\tau_{\mathrm{lim}}\triangleq0.85\,\bar{\tau}$ is the planning torque
limit.

Each resulting trajectory and its first two derivatives were evaluated
on the inclusive uniform grid
$t=n/F_s$, with $F_s=1000~\mathrm{Hz}$,
$n=0,\ldots,10000$, and $N_t=10001$. This grid defines the common
reference and diagnostic sampling instants.

Let $\bar{q}$, $\bar{\dot{q}}$, $\bar{\ddot{q}}$, and $\bar{\tau}$
denote the physical position, velocity, acceleration, and torque limit
vectors, respectively. The qualification bounds were defined as
\[
\begin{gathered}
q_{\min} = -0.90\,\bar{q},
\qquad
q_{\max} = 0.90\,\bar{q},
\\
\dot{q}_{\mathrm{lim}} = 0.85\,\bar{\dot{q}},
\quad
\ddot{q}_{\mathrm{lim}} = 0.85\,\bar{\ddot{q}},
\quad
\tau_{\mathrm{lim}} = 0.85\,\bar{\tau}.
\end{gathered}
\]
A candidate trajectory was accepted only if all evaluated quantities
were finite and, for every $t$,
\begin{equation}
\label{eq:full_grid_qualification}
\begin{gathered}
q_{\min} \preceq q_{\mathrm{ref}}(t) \preceq q_{\max}, \\
\left|\dot{q}_{\mathrm{ref}}(t)\right| \preceq \dot{q}_{\mathrm{lim}}, \quad
\left|\ddot{q}_{\mathrm{ref}}(t)\right| \preceq \ddot{q}_{\mathrm{lim}}, \quad
\left|\tau_{\mathrm{nom}}(t)\right| \preceq \tau_{\mathrm{lim}}.
\end{gathered}
\end{equation}
The nominal torque $\tau_{\mathrm{nom}}$ was evaluated using the
inverse-dynamics mapping defined in
Section~\ref{subsubsec:nominal_mapping}.

An empirical Joint~2 activity filter was additionally imposed:
\begin{equation}
\label{eq:joint2_activity_filter}
L_{q_2}
\triangleq
\int_{0}^{T}
\left|
\dot{q}_{\mathrm{ref},2}(t)
\right|
\,dt
\geq
0.8~\mathrm{rad}.
\end{equation}
The integral was evaluated by the trapezoidal rule on the same uniform
grid, excluding trajectories with negligible Joint~2 motion.

Before physical replay, each qualified trajectory and its
qualification record were frozen and reused identically across all
physical runs.

\subsection{Closed-Loop Physical System and Physical Campaign}
\label{subsec:physical_system}

\subsubsection{Constrained Flatness-MPC and Nominal Torque Mapping}
\label{subsubsec:mpc_torque_mapping}

To track the planned joint trajectory subject to simulation limits
adapted from the manufacturer specifications~\cite{kinova2022guide},
the controller combines a constrained flatness-based MPC, following
the general approach of~\cite{Wang2020FlatnessMPC}, with the nominal
manipulator inverse-dynamics mapping~\cite{lynch2017modern}. The MPC
computes a virtual-acceleration correction
$\Delta v_{k|k}\in\mathbb{R}^{6}$ to the planned acceleration. Its
internal prediction model consists of six double integrators
discretized with prediction interval $T_p=0.05~\mathrm{s}$. The Simulink controller block executes with inherited continuous sample time. The interval $T_p$ defines the prediction grid and reference indexing.

Let
\begin{equation}
\label{eq:mpc_decision_vector}
\Delta V_k
\triangleq
\begin{bmatrix}
\Delta v_{k|k}^{\top} &
\cdots &
\Delta v_{k+N_{c,k}-1|k}^{\top}
\end{bmatrix}^{\top}
\end{equation}
denote the sequence of acceleration corrections, and let
$e_{q,k+j|k}\triangleq q_{k+j|k}-q_{\mathrm{ref}}(t_{k+j})$ denote the
predicted joint-position tracking error. At each evaluation instant,
the correction sequence is obtained from the quadratic program (QP)
\begin{equation}
\label{eq:mpc_objective}
\begin{aligned}
\Delta V_k^{\star}
=
\arg\min_{\Delta V_k}\;
&20\sum_{j=1}^{N_{p,k}}
\left\|
e_{q,k+j|k}
\right\|_2^2
+
0.1
\left\|
\Delta v_{k|k}
\right\|_2^2
\\
&+
0.5\sum_{j=1}^{N_{c,k}-1}
\left\|
\Delta v_{k+j|k}
-
\Delta v_{k+j-1|k}
\right\|_2^2
\\
&+
10^{-3}
\left\|
\Delta V_k
\right\|_2^2.
\end{aligned}
\end{equation}
The QP is subject to the predicted position, velocity, acceleration, and
nominal-torque limits summarized in
Table~\ref{tab:mpc_parameters}. The maximum prediction and control
horizons are shortened near the end of the finite reference according
to
\begin{equation}
\label{eq:mpc_shrinking_horizons}
N_{p,k}
=
\min\left(15,N_{\mathrm{ref}}-k\right),
\qquad
N_{c,k}
=
\min\left(10,N_{p,k}\right).
\end{equation}
Here, $N_{\mathrm{ref}}$ is the number of samples on the internal
prediction grid. Table~\ref{tab:mpc_parameters} reports the implemented
simulation bounds, including a chosen interval of $[-\pi,\pi]$ for
the continuously rotating joints. The acceleration and nominal-torque
bounds adopt the manufacturer's hard acceleration and soft torque limits,
respectively~\cite[pp.~98--99]{kinova2022guide}.

\begin{table}[t]
\caption{Flatness-MPC and Torque-Mapping Parameters}
\label{tab:mpc_parameters}
\centering
\footnotesize
\renewcommand{\arraystretch}{1.12}
\begin{tabular}{@{} p{0.36\columnwidth} p{0.58\columnwidth} @{}}
\toprule
\textbf{Parameter} & \textbf{Implemented Value} \\
\midrule
Prediction interval & $T_p=0.05~\mathrm{s}$ \\
Maximum horizons & $N_{p,\max}=15$, $N_{c,\max}=10$ \\
Joint-position bounds & Joints 1, 4, 6: $\pm\pi~\mathrm{rad}$; \newline Joints 2, 3, 5: $\pm[2.25,\,2.58,\,2.10]^{\top}~\mathrm{rad}$ \\
Joint-velocity bounds & $\pm1.3963~\mathrm{rad/s}$ (Joints 1--3); \newline $\pm1.2217~\mathrm{rad/s}$ (Joints 4--6) \\
Virtual-acceleration bounds & $\pm5.2~\mathrm{rad/s^2}$ (Joints 1--3); \newline $\pm10~\mathrm{rad/s^2}$ (Joints 4--6) \\
Nominal-torque bounds & $\pm39~\mathrm{N\cdot m}$ (Joints 1--3); \newline $\pm9~\mathrm{N\cdot m}$ (Joints 4--6) \\
QP solver & \texttt{quadprog}, active-set algorithm \\
\bottomrule
\end{tabular}
\end{table}

The MPC formulation incorporates no separate terminal cost or invariant set, and operates without auxiliary feedback gains ($K_p=K_d=0$). The resulting virtual acceleration is
\begin{equation}
\label{eq:mpc_virtual_acceleration}
v_{\mathrm{total}}(t)
=
\begin{cases}
\ddot{q}_{\mathrm{ref}}(t_k)
+
\Delta v_{k|k}^{\star},
& \text{if the QP converges},
\\[1mm]
\ddot{q}_{\mathrm{ref}}(t_k),
& \text{otherwise}
\end{cases}.
\end{equation}
Here, the second case is the implemented feedforward fallback. The
commanded joint torque is then evaluated by mapping the actual joint-state
feedback and the MPC-corrected acceleration through the nominal
inverse-dynamics mapping~\eqref{eq:nominal_torque_ph}:
\begin{equation}
\label{eq:computed_torque_mapping}
\tau_{\mathrm{cmd}}(t)
=
\tau_{\mathrm{nom}}\!\left(q(t), \dot{q}(t), v_{\mathrm{total}}(t)\right).
\end{equation}
This mapping evaluates the nominal momentum dynamics~\eqref{eq:nominal_torque_ph}
using the measured joint state $(q(t), \dot{q}(t))$ and the MPC-corrected
acceleration $v_{\mathrm{total}}(t)$ in place of the planned reference
$\left(q_{\mathrm{ref}},\dot{q}_{\mathrm{ref}},\ddot{q}_{\mathrm{ref}}\right)$.
In contrast, $\tau_{\mathrm{nom}}(t)$ is evaluated exclusively along the
planned reference trajectory.

\subsubsection{Perturbed-Plant Dynamics and Joint-2 Anomaly Laws}
\label{subsubsec:perturbed_plant_faults}

The continuous physical system is the Kinova Gen3 instantiation of the
perturbed port-Hamiltonian model in~\eqref{eq:faulty_ph}, with
$u_{\mathrm{cmd}}=\tau_{\mathrm{cmd}}$ and $y=\dot{q}$. The nominal
reference generator and controller use the nominal model, whereas the
physical simulations use fixed inertial and dissipative perturbations.
The nominal link masses, centers of mass, and inertia tensors match the
manufacturer's six-DoF configuration without the vision module
\cite[pp.~207--209]{kinova2022guide}.

The nominal viscous and Coulomb coefficients were adapted from the
actuator-type identification in~\cite[Sec.~VII-A, eq.~(12)]{muraccioli2025demonstrating}.

For $i\in\{1,\ldots,6\}$, the plant parameters were defined by
\begin{equation}
\label{eq:plant_parameter_mismatch}
\begin{aligned}
m_i
&=1.05\,m_{i,\mathrm{nom}},
\\
c_i
&=
c_{i,\mathrm{nom}}
+
\begin{bmatrix}
0.002 & -0.002 & 0.002
\end{bmatrix}^{\top}\mathrm{m},
\\
I_{\mathrm{com},i}
&=
1.03\,I_{\mathrm{com},i,\mathrm{nom}},
\\
F_v
&=
\gamma_v\odot F_{v,\mathrm{nom}},
\qquad
F_c
=
\gamma_c\odot F_{c,\mathrm{nom}},
\\
F_{v,\mathrm{nom}}
&=
\begin{bmatrix}
2 & 2 & 2 & 1.36 & 1.36 & 1.36
\end{bmatrix}^{\top},
\\
\gamma_v
&=
\begin{bmatrix}
1.12 & 1.05 & 1.15 & 1.08 & 1.10 & 1.04
\end{bmatrix}^{\top},
\\
F_{c,\mathrm{nom}}
&=
\begin{bmatrix}
2.15 & 2.15 & 2.15 & 1.60 & 1.60 & 1.60
\end{bmatrix}^{\top},
\\
\gamma_c
&=
\begin{bmatrix}
1.08 & 1.06 & 1.10 & 1.05 & 1.09 & 1.03
\end{bmatrix}^{\top}.
\end{aligned}
\end{equation}
Here, $F_v$ and $F_c$ are expressed in
newton meter seconds per radian and newton meters, respectively. All
perturbations were deterministic, common to every physical run, and
constant throughout each trajectory.

The implemented friction law was
$\tau_f=D(\dot{q})\dot{q}$, with
$D(\dot{q})=\operatorname{diag}(D_1,\ldots,D_6)$ defined as
\begin{equation}
\label{eq:plant_dissipation_coefficient}
D_i(\dot{q}_i)
=
\begin{cases}
\tanh\!\left(\kappa|\dot{q}_i|\right)
\left[
F_{v,i}
+
F_{c,i}
\dfrac{\tanh(\kappa\dot{q}_i)}{\dot{q}_i}
\right],
& \dot{q}_i\neq0,
\\[2mm]
0,
& \dot{q}_i=0
\end{cases}.
\end{equation}
Here, $\kappa=50~\mathrm{s/rad}$. This definition uses the continuous limit at $\dot{q}_i=0$ and
satisfies $D(\dot{q})\succeq0$.

All anomalies acted exclusively on Joint~2. Let
$e_2\in\mathbb{R}^{6}$ be the second canonical basis vector and define
$\sigma_f(t)\triangleq\mathbf{1}_{[t_f,\infty)}(t)$, with
$t_f\sim\mathcal{U}(2,5)~\mathrm{s}$. The three anomaly laws and the training and
validation parameter marginals were
\begin{subequations}
\label{eq:joint2_anomaly_laws}
\begin{align}
\Pi_{\mathrm{LoE}}(t)
&=
I_6-s_{\mathrm{LoE}}\sigma_f(t)e_2e_2^\top,
\label{eq:loe_anomaly}
\\[-0.5mm]
s_{\mathrm{LoE}}
&\sim\mathcal{U}(0.02,0.30),
\notag
\\[1mm]
\tau_{f,\mathrm{flt}}(t)
&=
e_2\sigma_f(t)\Delta F_c
\tanh\!\left(\kappa\dot q_2(t)\right),
\label{eq:friction_anomaly}
\\[-0.5mm]
\Delta F_c
&=
2.15s_F~\mathrm{N\cdot m},
\quad
s_F
\sim\mathcal{U}(0.02,1.50),
\notag
\\[1mm]
d_\tau(t)
&=
e_2\sigma_f(t)A_d
\sin\!\left(2\pi f_dt+\phi_d\right),
\label{eq:vibration_anomaly}
\\[-0.5mm]
A_d
&\sim\mathcal{U}(0.1,2.5)~\mathrm{N\cdot m},
\quad
f_d
\sim\mathcal{U}(2,30)~\mathrm{Hz},
\notag
\\[-0.5mm]
\phi_d
&\sim\mathcal{U}(0,2\pi).
\notag
\end{align}
\end{subequations}
These terms instantiate actuator loss of effectiveness
\cite{RobustFTC_PH}, additional Coulomb dissipation, and exogenous
harmonic effort~\cite{randall2022vibration}, respectively. The friction
torque in~\eqref{eq:friction_anomaly} realizes the additional
dissipative effort $\Delta D_{\mathrm{flt}}(x)\dot q$ of
Section~\ref{sec:physical_system} on Joint~2, activated here by
$\sigma_f(t)$. It enters the momentum dynamics as
$-\tau_{f,\mathrm{flt}}(t)$ and satisfies
$\dot q^\top(t)\tau_{f,\mathrm{flt}}(t)\ge0$.
The vibration phase was evaluated using the global time base $t$. Each
run contains at most one anomaly; the nominal condition
uses $\Pi=I_6$, $\tau_{f,\mathrm{flt}}=0$, and $d_\tau=0$.

\subsubsection{Paired Physical Campaign Design}
\label{subsubsec:paired_campaign}

The Primary campaign comprised $5000$ qualified parent trajectories,
each evaluated through four independent closed-loop siblings:
nominal, loss of effectiveness, anomalous friction, and vibration.
The siblings shared the same frozen reference trajectory and motion
mode, while the three faulty runs also shared the parent-specific onset
time $t_f$. Because each sibling was simulated independently, their
physical states and measured signals were allowed to diverge after
fault activation. This produced $20\,000$ Primary physical runs.

Parents were assigned to training, validation, and testing using a
$70/15/15$ grouped split, corresponding to $3500$, $750$, and $750$
parents, respectively. All four siblings of a parent remained in the
same partition, preventing reference-trajectory identity from crossing
between data splits. The segment counts $K\in\{2,3\}$ and the
\texttt{fly}/\texttt{stop} modes were balanced within each Primary
partition.

Training and validation used Latin hypercube sampling over the stated parameter ranges.
Primary test allocated 187--188 runs to each LoE and Friction severity bin and
 31--32 runs to each vibration amplitude--frequency cell used in E1 and E2-A.

An additional controlled frequency-sweep stratum comprised $30$
parents disjoint from Primary, with ten vibration runs per parent at
fixed amplitude $A_d=1~\mathrm{N\cdot m}$ and
$f_d\in\{2,5,8,11,14,17,20,23,26,30\}~\mathrm{Hz}$. Within each
family, the reference geometry was fixed while the vibration frequency
was varied. These $300$ runs formed an evaluation-only
frequency-sweep stratum.

The complete physical campaign therefore contained $5030$ parents and
$20\,300$ runs. Each run retained its parent, frozen trajectory, physical
condition, fault parameters, and grouped-partition identifiers.

\subsubsection{Numerical Integration and Signal Acquisition}
\label{subsubsec:numerical_acquisition}

The continuous closed-loop dynamics were integrated using MATLAB
\texttt{ode45} with
$\mathrm{RelTol}=10^{-6}$,
$\mathrm{AbsTol}=10^{-8}$, and
$\mathrm{MaxStep}=10^{-3}~\mathrm{s}$.
The solver retained adaptive internal steps, and the resulting
variable-step physical telemetry was resampled onto the $1~\mathrm{kHz}$
diagnostic grid using shape-preserving piecewise cubic Hermite
interpolation (\texttt{pchip}). The retained synchronized signals
comprised $q(t)$, $\dot q(t)$, and $\tau_{\mathrm{cmd}}(t)$, aligned
with $q_{\mathrm{ref}}(t)$, $\dot q_{\mathrm{ref}}(t)$,
$\ddot q_{\mathrm{ref}}(t)$, and $\tau_{\mathrm{nom}}(t)$.

For the physical campaign,
$q_{\mathrm{meas}}(t)=q(t)$ and
$\dot q_{\mathrm{meas}}(t)=\dot q(t)$. The sampled commanded-power
signal was evaluated as
\begin{equation}
\label{eq:measured_power_profile}
\mathcal{P}_{\mathrm{cmd},i}(t)
=
\dot q_{\mathrm{meas},i}(t)
\tau_{\mathrm{cmd},i}(t),
\qquad
i\in\{1,\ldots,6\}.
\end{equation}

Matched reruns compared the reported \texttt{ode45} configuration
with \texttt{MaxStep=auto}, holding the trajectory, physical condition,
fault parameters, and controller configuration fixed. Comparison on
the uniform resampling grid showed differences in commanded torque
and diagnostic residuals; the $1$-ms maximum-step configuration was
retained for the reported campaign.

\subsection{Spectro-Temporal Fault Diagnosis and Matched Diagnostic Comparators}
\label{subsec:spectro_temporal_dx}

\subsubsection{Residual Construction}
\label{subsubsec:residual_construction}

For the manipulator, the SHAPE residual
$r_{\mathrm{SHAPE},i}(t)$ defined in
\eqref{eq:residual_def} was evaluated for
$i\in\{1,\ldots,6\}$. For comparison, three established model-based residuals were implemented as diagnostic baselines. The torque residual adapted the inverse-dynamics comparison in~\cite{haddadin2008collision} by replacing the measured joint torque with the commanded joint torque; it was computed for each joint as the commanded-minus-nominal torque difference. The generalized momentum observer (GMO) followed the formulation in~\cite{deluca2003actuator}, using the measured joint positions and velocities, commanded joint torques, and nominal inertia, Coriolis, gravity, and friction terms to produce six joint-resolved residuals. The energy observer (EO) followed the kinetic-energy formulation in~\cite[Sec.~III-A, eq.~(28)]{haddadin2017robot}, producing one scalar residual from the measured joint motion, commanded joint torques, and nominal inertia, gravity, and friction terms.

Both observers used the gain $k_o=100$ and were discretized through the bilinear Tustin transformation on the diagnostic time grid defined in Section~\ref{subsubsec:numerical_acquisition}. Their internal states were initialized from the corresponding nominal momentum and kinetic energy at $t=0$, yielding zero initial residuals. The resulting residual bank comprised six joint-resolved SHAPE residuals, six torque residuals, six GMO residuals, and one scalar EO residual.

\subsubsection{Physical Scaling and Spectro-Temporal Representation}
\label{subsubsec:cwt_discretization}

To realize the abstract spectro-temporal mapping $\mathcal{T}_{\mathrm{ST}}$ formulated in Section~\ref{sec:tfa_extraction}, each residual family was first rendered nondimensional using fixed physical scaling constants common to all trajectories, for $i\in\{1,\ldots,6\}$:
\begin{alignat}{2}
\tilde{r}_{\mathrm{SHAPE},i}(t)
&=\frac{r_{\mathrm{SHAPE},i}(t)}{\bar{\tau}_i\bar{\dot{q}}_i},
&\quad\tilde{r}_{\tau,i}(t)
&=\frac{r_{\tau,i}(t)}{\bar{\tau}_i},\notag\\[1mm]
\tilde{r}_{\mathrm{GMO},i}(t)
&=\frac{r_{\mathrm{GMO},i}(t)}{\bar{\tau}_i},
&\quad\tilde{r}_{\mathrm{EO}}(t)
&=\frac{r_{\mathrm{EO}}(t)}{\sum_{j=1}^{6}\bar{\tau}_j\bar{\dot{q}}_j}.
\label{eq:physical_normalization}
\end{alignat}
Here, $\bar{\tau}_i$ and $\bar{\dot{q}}_i$ denote the actuator torque and joint velocity limits reported in Table~\ref{tab:mpc_parameters}.

The scaled residual channels were then transformed using the continuous wavelet transform (CWT) as the concrete time--frequency operator. For any normalized scalar residual channel $\tilde r(t)$, the transform was evaluated under the $L^1$ normalization convention of the implemented filterbank~\cite{mathworksCwtFilterbank}:
\begin{equation}
\label{eq:cwt_l1_definition}
W_{\tilde r}(a,b)
=
\frac{1}{a}
\int_{-\infty}^{\infty}
\tilde r(t)
\psi^{*}\!\left(\frac{t-b}{a}\right)
\,dt,
\qquad
a>0;\quad b\in\mathbb{R}.
\end{equation}
Here, $a$ is scale, $b$ is temporal translation, and $\psi$ is an admissible analytic mother wavelet satisfying $C_{\psi} = \int_{0}^{\infty} |\widehat{\psi}(\omega)|^{2}\omega^{-1}\,d\omega < \infty$~\cite{mallat1999wavelet}. The factor $a^{-1}$ matches the $L^1$ normalization used by the implemented filterbank.

Computationally, the transform was instantiated in MATLAB R2026a using an analytic Morlet wavelet filterbank (\texttt{amor}) with 32~voices per octave over the physical bandwidth $[1, 100]~\mathrm{Hz}$. Each channel was processed independently using the native center-frequency and temporal coordinates returned by the filterbank.

To construct standardized discrete inputs, native coefficients were aggregated onto a grid of $N_\lambda=224$~frequency bins and $N_b=448$~temporal bins via overlap-weighted aggregation in physical frequency--time coordinates. Coefficient validity was determined from the native cone of influence (COI), with a coefficient at frequency $f$ and time $t$ treated as valid when $f \ge f_{\mathrm{COI}}(t)$. Native coefficients outside this valid domain were excluded from aggregation. For each destination grid cell $(m,n)$, the valid support fraction $\gamma_{mn}\in[0,1]$ was computed as the ratio of valid native area to total bin area, and the cell was accepted if $\gamma_{mn} \ge 0.80$.

The primary representation $[\mathbf{X}_i]_{mn}=Z_{i,mn}$ was converted to logarithmic power. For accepted cells,
\begin{equation}
\label{eq:log_power_cwt}
Z_{i,mn}=\max\bigl(10\log_{10}(\bar{P}_{i,mn}+10^{-10}),-100\bigr).
\end{equation}
Cells with $\gamma_{mn}<0.80$ were assigned $Z_{i,mn}=-100$.
Here, $i$ indexes the transformed residual component, $m\in\{1,\ldots,N_\lambda\}$ indexes the spectral bin, $n\in\{1,\ldots,N_b\}$ indexes the temporal bin, and $\bar{P}_{i,mn}$ denotes the overlap-weighted CWT power of component $i$ computed from valid native support. For each trajectory, the SHAPE, torque, and GMO families produced six maps each, while the energy observer produced one map. All maps were stored as $224 \times 448$ single-precision (\texttt{float32}) arrays.
\subsubsection{Matched Diagnostic Input Configurations}
\label{subsubsec:diagnostic_configurations}

Two additional input representations were constructed from the physically scaled SHAPE residuals. The temporal configuration retained the six joint sequences over the $N_t=10\,001$ samples of the diagnostic grid. For the fast Fourier transform (FFT) configuration, each joint sequence was multiplied by a symmetric Hann window of length $N=10\,001$. An $N$-point discrete Fourier transform was then evaluated without mean subtraction or zero-padding. Amplitudes were normalized by the sum of the window samples, and the nonzero-frequency amplitudes in the retained range were doubled according to the single-sided amplitude convention. The first $1001$ nonnegative-frequency bins were retained, with $f_k=kF_s/N$ for $k\in\{0,\ldots,1000\}$, corresponding to frequencies from zero to approximately $100~\mathrm{Hz}$. The resulting six diagnostic input configurations are summarized in Table~\ref{tab:diagnostic_configurations}.

\begin{table}[t]
\centering
\caption{Diagnostic Input Configurations}
\label{tab:diagnostic_configurations}
\footnotesize
\begin{tabular}{@{}lll@{}}
\toprule
Configuration & Representation & Input space \\
\midrule
M01: SHAPE--CWT      & 2-D CWT log-power & $\mathbb{R}^{6\times224\times448}$ \\
M02: Torque--CWT     & 2-D CWT log-power & $\mathbb{R}^{6\times224\times448}$ \\
M03: GMO--CWT        & 2-D CWT log-power & $\mathbb{R}^{6\times224\times448}$ \\
M04: EO--CWT         & 2-D CWT log-power & $\mathbb{R}^{1\times224\times448}$ \\
M05: SHAPE--Temporal & 1-D time sequence & $\mathbb{R}^{6\times10\,001}$ \\
M06: SHAPE--FFT      & 1-D amplitude spectrum & $\mathbb{R}^{6\times1001}$ \\
\bottomrule
\end{tabular}
\end{table}

Configurations M01--M04 compare the four residual families under the common CWT construction defined in Section~\ref{subsubsec:cwt_discretization}. Configurations M01, M05, and M06 compare the CWT, temporal, and Fourier representations derived from the same physically scaled SHAPE residuals. All configurations used the parent-grouped data partition defined in Section~\ref{subsubsec:paired_campaign}.

\subsubsection{Shared-Encoder Classifier Architectures}
\label{subsubsec:cnn_architecture}

\begin{figure*}[t]
    \centering
    \includegraphics[
        width=\textwidth
    ]{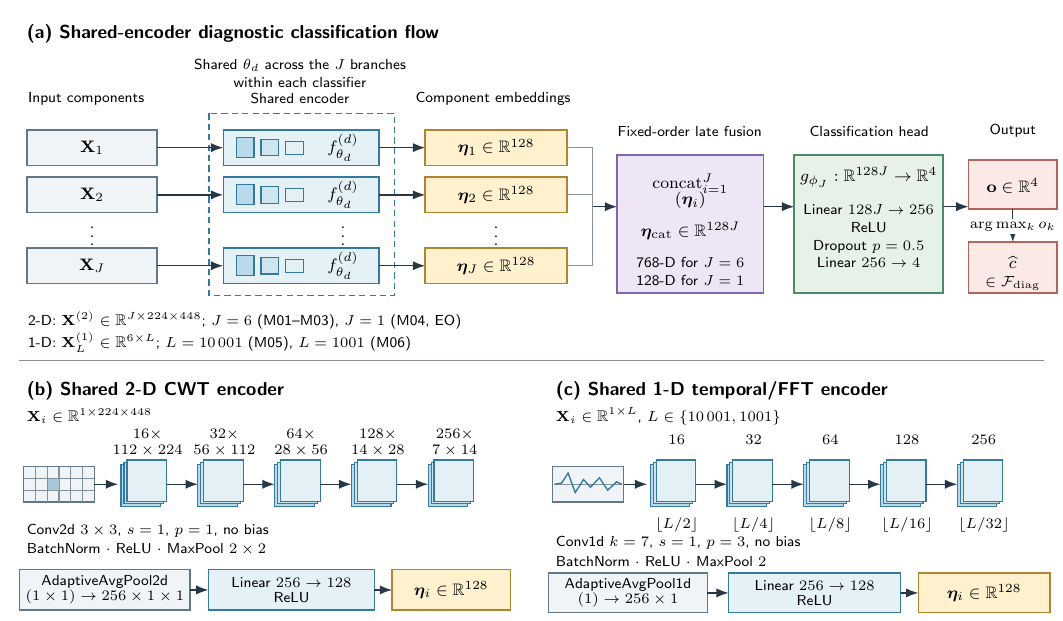}
    \caption{Shared-encoder classifier architectures used for the six diagnostic configurations. (a) Component-wise encoding, fixed-order late fusion, and four-class classification flow. (b) Two-dimensional encoder used for the CWT configurations. (c) One-dimensional encoder used for the temporal and Fourier configurations. Encoder parameters are shared across the active branches within each independently trained classifier.}
    \label{fig:shared_encoder_diagnostic_architectures}
\end{figure*}

The primary SHAPE--CWT configuration (M01) directly instantiates the inductive classification operator $\mathcal{C}$ formulated in Section~\ref{sec:pattern_classification}, mapping the composite representation $\mathbf{X}$ to the discrete health classes $\mathcal{F}_{\mathrm{diag}}$. The comparator configurations (M02--M06) are classified through analogous shared-encoder architectures adapted to their respective input spaces (Table~\ref{tab:diagnostic_configurations}). The continuous wavelet representations (M01--M04) are processed by a two-dimensional encoder, whereas the temporal sequence (M05) and Fourier amplitude spectrum (M06) are processed by a one-dimensional encoder. Within each classifier, the encoder parameters are shared across its active branches. Here, $J=6$ for M01--M03 and M05--M06, whereas $J=1$ for the single global EO component in M04. The common classification flow and the internal structures of both encoders are illustrated in Fig.~\ref{fig:shared_encoder_diagnostic_architectures}.

\paragraph{Shared encoders}
Each input matrix or sequence $\mathbf{X}_i$ is presented to its respective encoder with a singleton channel axis. The two architectural families map single-component inputs to a common 128-dimensional latent space:
\begin{equation}
\label{eq:shared_encoder_maps}
\begin{aligned}
f_{\theta_2}^{(2)}
&\colon
\mathbb{R}^{1\times224\times448}
\rightarrow
\mathbb{R}^{128},
\\
f_{\theta_1}^{(1)}
&\colon
\mathbb{R}^{1\times L}
\rightarrow
\mathbb{R}^{128},
\qquad
L\in\{10\,001,1001\}.
\end{aligned}
\end{equation}
These maps yield branch embeddings $\boldsymbol{\eta}_i \in \mathbb{R}^{128}$ for each input component $\mathbf{X}_i$ ($i\in\{1,\ldots,J\}$). The 2-D encoder comprises five sequential convolutional blocks ($3\times3$ convolutions, stride 1, padding 1, no bias, batch normalization, rectified linear unit (ReLU) activation, and $2\times2$ max-pooling) across channels $16, 32, 64, 128, 256$, followed by adaptive average pooling and a linear projection from 256 to 128 components. The 1-D encoder employs the same channel progression over five blocks (kernel 7, stride 1, padding 3, no bias, batch normalization, ReLU, and max-pooling of 2), with adaptive average pooling and a linear projection from 256 to 128 components.

\paragraph{Late fusion and classification}
Branch embeddings are aggregated through fixed-order concatenation:
\begin{equation}
\label{eq:feature_fusion}
\boldsymbol{\eta}_{\mathrm{cat}}
=
\operatorname{concat}_{i=1}^{J}
(\boldsymbol{\eta}_i)
\in
\mathbb{R}^{128J},
\qquad
J\in\{1,6\}.
\end{equation}
For $J=6$, embeddings are concatenated in fixed joint order.

The composite representation is mapped to class logits $\mathbf{o} \in \mathbb{R}^4$ and a predicted class $\widehat{c}$ by a multi-layer perceptron head:
\begin{equation}
\begin{aligned}
\mathbf{h}
&=
\operatorname{ReLU}
\left(
\mathbf{W}_1\boldsymbol{\eta}_{\mathrm{cat}}
+\mathbf{b}_1
\right)
\in\mathbb{R}^{256},
\\
\mathbf{o}
&=
\mathbf{W}_2
\operatorname{Dropout}_{0.5}(\mathbf{h})
+\mathbf{b}_2
\in\mathbb{R}^4,
\\
\widehat{c}
&=
\operatorname*{arg\,max}_{k\in\{1,\ldots,4\}}o_k
\in\mathcal{F}_{\mathrm{diag}}.
\end{aligned}
\end{equation}
Dropout is applied during training. Counting shared encoder parameters once, these specifications yield $623\,604$ trainable parameters for the six-branch 2-D configurations (M01--M03), $459\,764$ for the single-branch 2-D configuration (M04), and $536\,532$ for the six-branch 1-D configurations (M05--M06).

\subsubsection{Training Protocol and Model Selection}
\label{subsubsec:optimization}

Each diagnostic configuration (M01--M06) was trained independently by minimizing the unweighted categorical cross-entropy over the four classes in $\mathcal{F}_{\mathrm{diag}}$. Optimization used AdamW~\cite{loshchilov2019decoupled} with an initial learning rate of $10^{-4}$ and weight decay of $10^{-4}$. The learning rate followed a cosine annealing schedule configured over a maximum of 50 epochs with a terminal value of zero. Mini-batch sizes were 32 for M01--M03, 64 for M04 and M06, and 16 for M05.

Validation was performed after each epoch. Checkpoints were selected by validation macro-$F_1$, with balanced accuracy and lower validation loss used as successive tie-breakers. Training terminated after ten consecutive epochs without improvement in this criterion.

For each configuration, training was repeated from scratch using the five fixed seeds $\{1729, 2027, 4099, 7919, 104729\}$. The checkpoint selected independently for each seed was evaluated on the held-out test partition only after model selection.

\subsubsection{Evaluation Protocols and Statistical Analysis}
\label{subsubsec:evaluation_protocols}

The trained diagnostic classifiers were evaluated through the six protocols summarized in Table~\ref{tab:evaluation_protocols}.

\begin{table}[t]
\centering
\caption{Diagnostic Evaluation Protocols}
\label{tab:evaluation_protocols}
\footnotesize
\begin{tabular}{@{}lp{0.78\columnwidth}@{}}
\toprule
Protocol & Scope and Evaluation Design \\
\midrule
E0 & Primary test benchmark ($3000$ runs, $750$ parents); M01--M06. \\
E1 & Primary fault runs stratified by physical anomaly severity; M01--M06; descriptive. \\
E2-A & Observational $4\times6$ amplitude--frequency grid on the $750$ Primary vibration runs; M01--M06; descriptive. \\
E2-B & Controlled $300$-run frequency sweep at fixed vibration amplitude; M01--M04; frequency-resolved recall intervals. \\
E3 & Primary test stratified by post-fault dynamic excitation quartiles; M01--M06; descriptive. \\
E4 & Ten offline sensor-perturbation conditions applied to Primary test trajectories; M01--M04; descriptive. \\
\bottomrule
\end{tabular}
\end{table}

For E1, physical anomaly severity was partitioned into four ascending bins, labeled B1 through B4 within each fault class:
\[
\begin{aligned}
s_{\mathrm{LoE}} &\colon \{[0.02,0.05), [0.05,0.10), [0.10,0.20), [0.20,0.30]\}, \\
s_F &\colon \{[0.02,0.10), [0.10,0.30), [0.30,0.75), [0.75,1.50]\}, \\
A_d &\colon \{[0.10,0.50), [0.50,1.00), [1.00,1.75), [1.75,2.50]\}.
\end{aligned}
\]
The dimensionless friction parameter $s_F$ corresponds to torque increments $\Delta F_c = 2.15s_F~\mathrm{N\cdot m}$~\eqref{eq:joint2_anomaly_laws}, whereas the $A_d$ intervals are expressed in newton meters. In E2-A, these four amplitude intervals were crossed with six frequency bands:
\[
[2,5),\ [5,10),\ [10,15),\ [15,20),\ [20,25),\ [25,30]~\mathrm{Hz}.
\]
These define a $24$-cell observational grid containing $31$--$32$ physical runs per cell. Protocol E2-B utilized the independent, fixed-amplitude vibration sweep defined in Section~\ref{subsubsec:paired_campaign}.

For E3, post-fault dynamic excitation was quantified from the root mean square (RMS) of the measured Joint-2 velocity:
\begin{equation}
\label{eq:post_fault_excitation}
\begin{aligned}
\dot q_{2,\mathrm{rms}}^{\mathrm{post}}
&=
\left(
\frac{1}{N_{\mathrm{post}}}
\sum_{k:t_k\geq t_f}
\dot q_{\mathrm{meas},2}^{\,2}(t_k)
\right)^{1/2},
\\
N_{\mathrm{post}}
&=
\sum_{k:t_k\geq t_f}1.
\end{aligned}
\end{equation}
The post-fault window comprised all samples satisfying $t_k\geq t_f$ through the terminal sample, and the recorded velocity was used directly without demeaning. Computational quartile thresholds were estimated exclusively from Primary training runs:
\[
\begin{aligned}
Q_{25} &= 0.1877854322~\mathrm{rad/s},\\
Q_{50} &= 0.2878207450~\mathrm{rad/s},\\
Q_{75} &= 0.4027505066~\mathrm{rad/s}.
\end{aligned}
\]
These thresholds partition $\dot q_{2,\mathrm{rms}}^{\mathrm{post}}$ into four evaluation strata:
\[
(-\infty,Q_{25}],\qquad
(Q_{25},Q_{50}],\qquad
(Q_{50},Q_{75}],\qquad
(Q_{75},\infty).
\]
The strata contain $782$, $676$, $774$, and $768$ runs, respectively.

For E4, sensor noise was simulated by perturbing the recorded joint telemetry, with velocity-noise level specified by the signal-to-noise ratio (SNR):
\begin{equation}
\label{eq:diagnostic_sensor_perturbations}
\begin{aligned}
q'_{j,k}
&=
q_{\mathrm{meas},j}(t_k)
+
\frac{\pi}{180}\sigma_q Z^q_{j,k},
\\
\dot q'_{j,k}
&=
\dot q_{\mathrm{meas},j}(t_k)
+
R_j 10^{-\mathrm{SNR}/20} Z^v_{j,k}.
\end{aligned}
\end{equation}
Here, $Z^q_{j,k}, Z^v_{j,k} \sim \mathcal{N}(0,1)$ are independent standard normal variables sampled per joint $j \in \{1,\ldots,6\}$ and time step $k$, and $R_j$ is the RMS velocity of joint $j$ pooled over all samples from the $14\,000$ Primary training runs. Ten perturbation conditions were evaluated: velocity noise alone at $\mathrm{SNR} \in \{40, 30, 20, 10\}~\mathrm{dB}$, position noise alone at $\sigma_q \in \{0.01, 0.05, 0.20\}^{\circ}$, and three combined conditions pairing $40~\mathrm{dB}$ velocity noise with each position-noise level. Position and velocity perturbations were applied offline to measurements while planned-reference and recorded controller quantities remained fixed; diagnostic residuals were then recomputed. The unperturbed E0 benchmark provided the clean reference.

Reported metrics comprise classification accuracy, macro-$F_1$, per-class recall, binary fault sensitivity (the fraction of faulty runs assigned to any fault class), missed-detection rate (MDR, the fraction of faulty runs classified as nominal), and false-alarm rate (FAR, the fraction of nominal runs classified as faulty). Per-class recall required identifying the specific ground-truth fault category. Metrics were evaluated separately for each of the five fixed training seeds and summarized by their sample mean and standard deviation.

For E2-B, parent-population uncertainty in per-frequency recall was evaluated via cluster resampling~\cite{field2007cluster} over the $30$ parents ($B=10\,000$, random-number-generator seed 20260927). For each frequency and method, a common resampled parent sequence was used across all five seeds, and the five seed-specific recalls were averaged within each replicate to obtain percentile $95\%$ intervals. These intervals describe parent-population uncertainty conditional on the frozen models. E0 metrics and evaluations under E1, E2-A, E3, and E4 are reported descriptively.

Code, archived results, and reproduction requirements are described in
the \hyperref[sec:data_availability]{Data Availability} section.


\section{Results and Discussion}
\label{sec:results}

\subsection{Clean Diagnostic Profiles}
\label{subsec:clean_diagnostic_profiles}

Under the common CWT representation, SHAPE achieves higher mean recall for all three fault classes and lower mean MDR than Torque, accompanied by higher mean FAR (Table~\ref{tab:e0_clean_profiles}). SHAPE and EO attain similar mean accuracies but different diagnostic profiles: SHAPE provides higher LoE recall, whereas EO yields higher Friction recall and lower MDR and FAR. GMO achieves the highest mean accuracy, macro-$F_1$, and binary fault sensitivity, together with the lowest MDR and FAR. These aggregate results establish the diagnostic performance of the power residuals relative to the alternative formulations; the following analysis examines how detection and fault-type identification vary with severity and operating excitation.

\suppressfloats[t]
\begin{table}[t]
\centering
\caption{Clean Primary Test Results Under CWT, Reported as Five-Seed Mean $\pm$ Sample Standard Deviation. Metrics Are Percentages Except Dimensionless Macro-$F_1$. FAR and MDR Denote False-Alarm and Missed-Detection Rates}
\label{tab:e0_clean_profiles}
\footnotesize
\setlength{\tabcolsep}{4pt}
\renewcommand{\arraystretch}{1.1}
\begin{tabular}{@{}lcccc@{}}
\toprule
Metric & SHAPE & Torque & GMO & EO\\
\midrule
\multicolumn{5}{@{}l}{\textit{(a) Overall classification and binary detection}}\\
\midrule
Accuracy & \begin{tabular}[c]{@{}c@{}} $84.69$ \\ $\pm0.70$ \end{tabular}
         & \begin{tabular}[c]{@{}c@{}} $81.67$ \\ $\pm0.70$ \end{tabular}
         & \begin{tabular}[c]{@{}c@{}} $97.17$ \\ $\pm0.37$ \end{tabular}
         & \begin{tabular}[c]{@{}c@{}} $84.62$ \\ $\pm1.70$ \end{tabular} \\
\addlinespace[2pt]
Macro-$F_1$ & \begin{tabular}[c]{@{}c@{}} $0.8453$ \\ $\pm0.0071$ \end{tabular}
            & \begin{tabular}[c]{@{}c@{}} $0.8166$ \\ $\pm0.0061$ \end{tabular}
            & \begin{tabular}[c]{@{}c@{}} $0.9716$ \\ $\pm0.0038$ \end{tabular}
            & \begin{tabular}[c]{@{}c@{}} $0.8433$ \\ $\pm0.0179$ \end{tabular} \\
\addlinespace[2pt]
Sensitivity & \begin{tabular}[c]{@{}c@{}} $88.05$ \\ $\pm1.65$ \end{tabular}
            & \begin{tabular}[c]{@{}c@{}} $83.77$ \\ $\pm1.92$ \end{tabular}
            & \begin{tabular}[c]{@{}c@{}} $96.95$ \\ $\pm0.35$ \end{tabular}
            & \begin{tabular}[c]{@{}c@{}} $90.54$ \\ $\pm1.22$ \end{tabular} \\
\addlinespace[2pt]
MDR & \begin{tabular}[c]{@{}c@{}} $11.95$ \\ $\pm1.65$ \end{tabular}
    & \begin{tabular}[c]{@{}c@{}} $16.23$ \\ $\pm1.92$ \end{tabular}
    & \begin{tabular}[c]{@{}c@{}} $3.05$ \\ $\pm0.35$ \end{tabular}
    & \begin{tabular}[c]{@{}c@{}} $9.46$ \\ $\pm1.22$ \end{tabular} \\
\addlinespace[2pt]
FAR & \begin{tabular}[c]{@{}c@{}} $6.21$ \\ $\pm1.71$ \end{tabular}
    & \begin{tabular}[c]{@{}c@{}} $5.33$ \\ $\pm1.98$ \end{tabular}
    & \begin{tabular}[c]{@{}c@{}} $0.19$ \\ $\pm0.07$ \end{tabular}
    & \begin{tabular}[c]{@{}c@{}} $2.24$ \\ $\pm0.56$ \end{tabular} \\
\midrule
\multicolumn{5}{@{}l}{\textit{(b) Multiclass recall}}\\
\midrule
Nominal & \begin{tabular}[c]{@{}c@{}} $93.79$ \\ $\pm1.71$ \end{tabular}
        & \begin{tabular}[c]{@{}c@{}} $94.67$ \\ $\pm1.98$ \end{tabular}
        & \begin{tabular}[c]{@{}c@{}} $99.81$ \\ $\pm0.07$ \end{tabular}
        & \begin{tabular}[c]{@{}c@{}} $97.76$ \\ $\pm0.56$ \end{tabular} \\
\addlinespace[2pt]
LoE & \begin{tabular}[c]{@{}c@{}} $79.20$ \\ $\pm2.62$ \end{tabular}
    & \begin{tabular}[c]{@{}c@{}} $72.88$ \\ $\pm4.84$ \end{tabular}
    & \begin{tabular}[c]{@{}c@{}} $98.43$ \\ $\pm0.30$ \end{tabular}
    & \begin{tabular}[c]{@{}c@{}} $70.19$ \\ $\pm5.88$ \end{tabular} \\
\addlinespace[2pt]
Friction & \begin{tabular}[c]{@{}c@{}} $66.45$ \\ $\pm2.18$ \end{tabular}
         & \begin{tabular}[c]{@{}c@{}} $64.05$ \\ $\pm4.53$ \end{tabular}
         & \begin{tabular}[c]{@{}c@{}} $90.43$ \\ $\pm1.58$ \end{tabular}
         & \begin{tabular}[c]{@{}c@{}} $71.07$ \\ $\pm2.18$ \end{tabular} \\
\addlinespace[2pt]
Vibration & \begin{tabular}[c]{@{}c@{}} $99.33$ \\ $\pm0.38$ \end{tabular}
          & \begin{tabular}[c]{@{}c@{}} $95.09$ \\ $\pm1.34$ \end{tabular}
          & \begin{tabular}[c]{@{}c@{}} $100.00$ \\ $\pm0.00$ \end{tabular}
          & \begin{tabular}[c]{@{}c@{}} $99.47$ \\ $\pm0.25$ \end{tabular} \\
\bottomrule
\end{tabular}
\end{table}

\subsection{Diagnostic Profiles Across Fault Severity and Operating Excitation}
\label{subsec:severity_excitation_profiles}

\begin{figure}[t]
\centering
\includegraphics[width=\columnwidth]{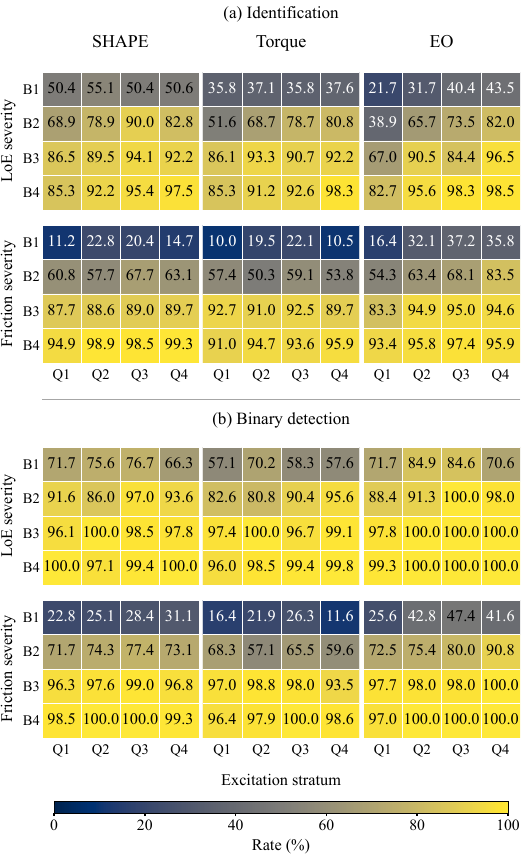}
\caption{LoE and Friction diagnosis by SHAPE, Torque, and EO under CWT on Primary test. Cells show five-seed mean percentages: (a) correct fault-type identification; (b) true faulty runs assigned to any fault class. B1--B4 follow the fault-specific severity intervals in Section~\ref{sec:simulation_setup}, from lowest to highest. Q1--Q4 are training-defined strata of post-fault measured Joint-2 velocity RMS. All panels share a 0\%--100\% scale.}
\label{fig:e3_joint_severity_excitation}
\end{figure}

Fig.~\ref{fig:e3_joint_severity_excitation} shows (a) correct fault-type identification and (b) binary fault sensitivity for SHAPE, Torque, and EO under CWT on Primary test. Q1--Q4 are the predefined strata of post-fault measured Joint-2 velocity RMS, with 29--81 unique simulated runs per severity--Q cell. Complete maps including GMO, exact cell counts, and between-seed sample standard deviations are available in the companion repository (see \hyperref[sec:data_availability]{Data Availability}).

LoE detection varies primarily with severity, with no consistent positive association between Q and binary sensitivity. In B1, sensitivity spans 66.27\%--76.67\% for SHAPE and 57.08\%--70.24\% for Torque; EO peaks at Q2 (84.88\%) before returning to 70.59\% at Q4. GMO remains near ceiling (96.86\%--99.17\%). Stronger severities bring the other formulations close to saturated detection. GMO's limited remaining variation near the ceiling prevents distinguishing weak Q dependence from saturation.

For weak LoE, SHAPE provides higher fault-type identification than Torque and EO throughout Q1--Q4. Its B1 identification remains within 50.42\%--55.12\%, whereas EO rises from 21.67\% at Q1 to 43.53\% at Q4. SHAPE thus combines higher identification with less variation across these strata than EO, while Torque also varies little but at a lower level (35.83\% at Q1). EO's higher binary sensitivity does not consistently translate into correct identification: at B1 Q2, it detects 84.88\% of faulty runs but identifies only 31.71\% as LoE, assigning 53.17\% to Friction. GMO retains substantially higher identification (91.37\%--97.56\% in B1).

At intermediate LoE severity, SHAPE's advantage over EO and Torque is largest at low Q and narrows toward Q4. In B2 Q1, identification is 68.95\% for SHAPE, 38.95\% for EO, and 51.58\% for Torque, despite similar SHAPE and EO binary sensitivities (91.58\% and 88.42\%). At Q4, identification converges to 82.80\%, 82.00\%, and 80.80\%, respectively. EO exceeds SHAPE in some stronger-fault, high-Q regions, including B3 Q4 (96.52\% versus 92.17\%). SHAPE also exhibits Q-associated identification gains: in B4, identification rises from 85.33\% at Q1 to 97.53\% at Q4 despite 100\% detection at both endpoints. Detectability and fault-type discrimination therefore remain distinct even for stronger LoE.

For the algebraic residuals, weak Friction is dominated by missed detections. For SHAPE in B1, binary sensitivity spans 22.80\%--31.05\% and identification 11.20\%--22.79\%. Torque has an irregular response, reaching only 11.58\% sensitivity and 10.53\% identification at Q4. EO's Q1-to-Q4 sensitivity rises from 25.60\% to 41.58\% (peaking at 47.37\% in Q3), with identification rising from 16.40\% to 35.79\%; its Q4 performance exceeds both algebraic formulations. GMO retains higher detection and identification, rising from 50.00\% to 78.42\% and from 48.40\% to 76.84\%, respectively. Thus, excitation-stratified differences in weak Friction affect whether faults are detected at all, as well as their assigned type.

For Friction B2, SHAPE exceeds Torque in both detection and identification throughout Q1--Q4. Its comparison with EO changes across strata: SHAPE has higher identification at Q1 (60.75\% versus 54.34\%), whereas EO reaches 83.46\% at Q4 versus 63.08\% for SHAPE. EO's binary sensitivity also rises from 72.45\% to 90.77\% between these endpoints. Its advantage at Q4 therefore accompanies improvements in both detection and identification, while SHAPE retains a smaller Q1-to-Q4 identification change.

The power residuals provide an identification advantage over Torque across weak and intermediate LoE and over EO in several low-Q regions, although weak-fault identification remains incomplete. EO closes or reverses this identification difference in specific higher Q regions. High identification is shared more broadly at strong Friction severity and strong LoE severity with high Q, where Torque also approaches SHAPE. GMO offers the broadest clean diagnostic coverage. These profiles distinguish regions of high identification from relative advantages within regions still dominated by confusion or missed detection.

These comparisons are conditioned on severity because its distribution varies across Q strata. Q1 denotes the lowest observed velocity-RMS stratum in the qualified trajectory population; post-fault velocity RMS covaries with severity, trajectory, and dynamics, so the profiles characterize observed operating regions rather than a controlled excitation effect.

\subsection{Amplitude--Frequency Dependence of Vibration Diagnosis}
\label{subsec:amplitude_frequency_vibration}

Vibration identification separates SHAPE from Torque in specific amplitude--frequency regions. In E2-A, at $A_d\in[0.10,0.50)~\mathrm{N\cdot m}$, Torque recall is $50.97\pm7.70\%$ at 15--20~Hz and $61.29\pm12.70\%$ at 20--25~Hz, while SHAPE, EO, and GMO achieve substantially higher recall in these bands. The power residuals thus support better discrimination than Torque alongside both observer formulations. Torque recovers to $96.77\%$ at 25--30~Hz, localizing its deficit rather than showing progressive deterioration with frequency.

The controlled E2-B sweep holds $A_d=1~\mathrm{N\cdot m}$ and reference geometry fixed within each parent family. All four formulations achieve $100\%$ observed identification at eight of ten frequencies, establishing shared high-performance regions. Torque reaches $88.67\pm3.80\%$ at 2~Hz and $92.00\pm5.58\%$ at 20~Hz. Its 20~Hz deficit therefore also occurs outside E2-A's lowest-amplitude interval. The errors have different diagnostic consequences: at 2~Hz, they are predominantly assignments to LoE ($10.67\%$), whereas at 20~Hz they are assignments to Nominal ($8.00\%$). The former preserves anomaly detection but confuses the fault type; the latter misses the anomaly. SHAPE's observed advantage over Torque is localized, while much of the frequency sweep offers little separation among the four formulations.

\subsection{Zero-Shot Diagnostic Response to Sensor Perturbations}
\label{subsec:zero_shot_sensor_perturbations}

E4 perturbs recorded position and velocity measurements offline and recomputes residuals under common CWT processing, with planned references and recorded controller quantities fixed. SHAPE's hard predictions remain unchanged under position-only perturbations, while Torque's remain unchanged across all ten conditions. Perturbed position does not enter SHAPE's final residual, and neither perturbed channel modifies Torque's residual in this evaluation.

Position perturbations alter both detection and identification for the observers. At $\sigma_q=0.01^\circ$, GMO LoE recall falls from $98.43\%$ to $44.40\%$, predominantly through assignments to Friction, while pooled MDR changes only from $3.05\%$ to $3.80\%$. Loss of LoE identification therefore coexists with little change in missed detection across the three faulty classes. For EO at the same noise level, the weakest LoE bin is frequently assigned to Friction, whereas the weakest Friction bin is predominantly assigned to Nominal. Stronger position perturbations increase EO's missed detections, with pooled MDR reaching $63.51\%$ at $0.20^\circ$. SHAPE retains its predictions across these position-only conditions, avoiding this direct measurement pathway for degradation.

Velocity perturbations expose a different limitation of SHAPE. At 40~dB, the weakest tested level, its predictions concentrate in Vibration: $98.59\%$ Vibration recall coexists with $98.77\%$ FAR, revealing loss of discrimination despite high class recall. GMO concentrates predictions in Nominal, with pooled MDR of $95.24\%$. EO exhibits both false alarms ($46.83\%$) and missed detections ($40.08\%$), followed by Vibration-directed predictions at stronger perturbations. At 20 and 10~dB, every GMO prediction is Nominal and every EO prediction is Vibration. Their identical $25\%$ accuracy on the balanced population thus represents opposite failures: missed faults versus indiscriminate alarms. Even in the strongest LoE bin, $s_{\mathrm{LoE}}\in[0.20,0.30]$, recall at 40~dB falls to $34.22\%$, $15.08\%$, and $1.28\%$ for SHAPE, GMO, and EO, respectively.

SHAPE's clean diagnostic gains over Torque therefore accompany additional exposure to velocity perturbations. Its final residual avoids the perturbed-position pathway affecting EO and GMO, but reliable discrimination still depends on velocity quality. The favorable clean operating regions must be interpreted alongside this sensitivity; the offline protocol characterizes the diagnostic measurement pathways with controller quantities fixed, leaving closed-loop sensor-noise tolerance unestablished.

\subsection{SHAPE Representation Dependence}
\label{subsec:shape_representation_dependence}

\begin{figure}[t]
\centering
\includegraphics[width=\columnwidth]{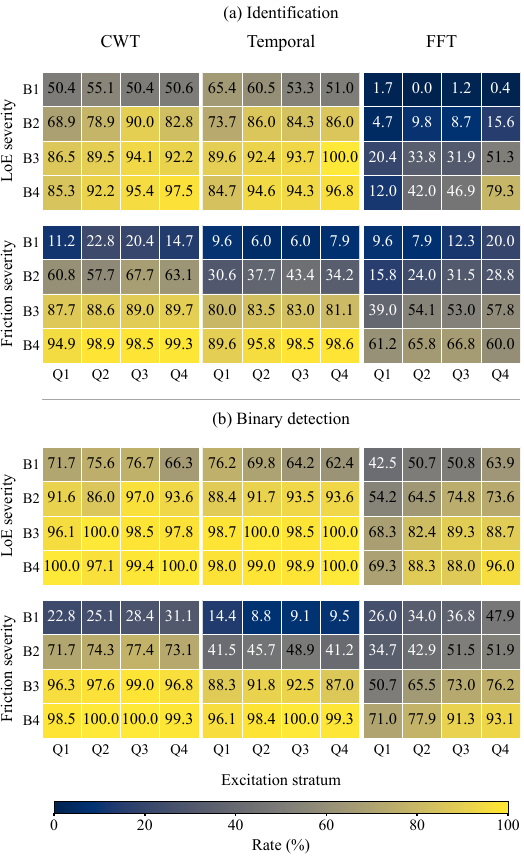}
\caption{LoE and Friction diagnosis by the CWT, Temporal, and FFT SHAPE pipelines on Primary test. Cells show five-seed mean percentages: (a) correct fault-type identification; (b) true faulty runs assigned to any fault class. B1--B4 follow the fault-specific severity intervals in Section~\ref{sec:simulation_setup}, from lowest to highest. Q1--Q4 are training-defined strata of post-fault measured Joint-2 velocity RMS. All panels share a 0\%--100\% scale.}
\label{fig:shape_joint_identification}
\end{figure}

The CWT, Temporal, and FFT comparisons concern complete SHAPE pipelines with different representations, architectures, dimensionalities, and training dynamics. On Primary test, mean accuracy is $84.69\pm0.70\%$, $82.37\pm0.75\%$, and $49.89\pm7.36\%$, respectively. Fig.~\ref{fig:shape_joint_identification} distinguishes fault-type identification from binary detection across severity--excitation strata. CWT's higher aggregate accuracy accompanies a broader Friction advantage, while its comparison with Temporal for LoE depends on the region and diagnostic metric.

In the weakest LoE bin, B1, CWT identification remains within $50.42\%$--$55.12\%$ across Q, while Temporal declines from $65.42\%$ at Q1 to $50.98\%$ at Q4. Temporal's initial advantage therefore narrows across the excitation strata, whereas CWT maintains a more constant, but still limited, identification profile. Temporal has higher identification and detection at Q1; at Q2--Q4, its higher identification instead coexists with more missed detections. For example, at Q3, Temporal identifies $53.33\%$ and detects $64.17\%$ of LoE runs, compared with $50.42\%$ and $76.67\%$ for CWT. CWT thus recognizes more faults in this cell but assigns more of them to another fault class. At stronger severities, both pipelines attain higher identification, with local reversals and non-monotonic Q trajectories rather than a uniform ordering.

Friction provides the clearest CWT--Temporal separation in B2: identification spans $57.71\%$--$67.66\%$ for CWT versus $30.57\%$--$43.40\%$ for Temporal, with higher CWT binary detection at every Q. CWT therefore improves both anomaly detection and fault-type identification. It also exceeds Temporal in both metrics throughout B1, although both pipelines remain strongly limited by missed detection. Their differences narrow in B4, where both form broad high-identification regions. The largest advantage at intermediate severity thus differs from the relative gains within the poorly detected weakest bin.

FFT exhibits substantial regional structure despite its lower aggregate performance and greater between-seed variability. In LoE B4, identification rises from $12.00\%$ at Q1 to $79.26\%$ at Q4, whereas identification in B1 remains very low despite appreciable binary detection. The favorable LoE regions are therefore concentrated at stronger severity and higher Q. Friction shows a different relationship between the metrics: in B4, FFT detection rises from $71.04\%$ to $93.10\%$ between Q1 and Q4, while identification changes from $61.19\%$ to $60.00\%$, with higher intermediate means. The higher observed detection at Q4 is accompanied by more wrong-fault assignments, rather than higher Friction identification. These profiles distinguish limited fault-type separability from failure to recognize abnormality.

Vibration identification is approximately $99.3\%$ for CWT and Temporal, compared with $71.41\%$ for FFT. Missed detections dominate FFT errors in every amplitude--Q cell, whereas wrong-fault assignments predominate for LoE. CWT's clearest benefit is in intermediate Friction; Temporal remains competitive in LoE and Vibration, with a detection--identification tradeoff in weak LoE. FFT's better LoE regions occur at stronger severity and higher Q, but remain below CWT and Temporal. These observational profiles associate each pipeline with different favorable regions and error patterns.

\subsection{Residual Evaluation Requirements}
\label{subsec:residual_evaluation_requirements}

With command torques and nominal references available, Torque requires one subtraction per joint, whereas SHAPE requires two multiplications and one subtraction (Table~\ref{tab:residual_evaluation_requirements}). Precomputing nominal power reduces SHAPE's final evaluation to one multiplication and one subtraction per joint: six of each per sample for $n_q=6$. Both algebraic mappings require $\mathcal{O}(n_q)$ arithmetic and maintain no recursive observer state. These counts exclude reference generation; both formulations retain nominal model-derived quantities.

GMO and EO evaluate nominal dynamics at the measured state and update recursive observer states. Both require inertia, gravity, and friction terms; GMO additionally requires $C^{\top}(q,\dot q)\dot q$, with $C(q,\dot q)\in\mathbb{R}^{n_q\times n_q}$ the nominal Coriolis--centrifugal matrix in the Christoffel convention. Given an assembled dense inertia matrix, momentum and kinetic-energy products require $\mathcal{O}(n_q^2)$ arithmetic, excluding model evaluation. The implemented central-difference routine for GMO evaluates inertia at twelve displaced configurations per sample, in addition to the base matrix. This count is specific to that six-joint implementation. GMO maintains six observer states and six joint-resolved residual channels; EO maintains one state and one scalar channel. SHAPE retains joint-resolved channels with measured velocity and nominal references, without measured-state dynamics evaluation. These requirements concern residual construction; complete-pipeline execution latency, including CWT and classification, remains unmeasured.

\begin{table}[t]
\centering
\caption{Six-Joint Residual-Evaluation Requirements. Inputs Concern the
Final Residual Mapping, Not Controller Sensing; All Methods Use Command
Torques. Reference Generation Is Excluded, and State Counts Denote
Recursive Observer Variables}
\label{tab:residual_evaluation_requirements}
\footnotesize
\setlength{\tabcolsep}{3pt}
\renewcommand{\arraystretch}{1.15}
\begin{tabular}{@{}lcccc@{}}
\toprule
Requirement & SHAPE & Torque & GMO & EO \\
\midrule
Measured position input & No & No & Yes & Yes \\
Measured velocity input & Yes & No & Yes & Yes \\
\shortstack[l]{Dynamics evaluated\\at measured state}
    & No & No & Yes & Yes \\
Observer states & 0 & 0 & 6 & 1 \\
Residual channels & 6 & 6 & 6 & 1 \\
\bottomrule
\end{tabular}
\end{table}

\subsection{Diagnostic Coverage and Conditional Tradeoffs}
\label{subsec:diagnostic_coverage_tradeoffs}

Algebraic port-power residuals support useful fault-type discrimination, with advantages over Torque concentrated in particular operating regions. Under common CWT processing, SHAPE improves identification across weak and intermediate LoE, improves both detection and identification in intermediate Friction, and avoids some localized Vibration deficits of Torque. Strong Friction and strong LoE at high Q also provide shared regions of high identification where the distinction narrows. Incorporating velocity into the algebraic construction therefore changes diagnostic coverage while retaining a simple final evaluation, but adds measurement exposure: in E4, velocity perturbations severely degrade SHAPE while leaving Torque's residual unchanged under the offline protocol.

Relative to EO, SHAPE has higher identification in several low-Q regions and less variation across Q in weak LoE. EO closes or reverses these differences in specific higher Q regions, so their similar aggregate accuracies conceal different operating profiles. SHAPE supplies joint-resolved algebraic residuals, whereas EO uses a scalar residual and a recursive state. GMO retains the broadest clean coverage, with the highest or tied highest detection and identification in every evaluated LoE--Friction severity--Q cell, alongside measured-state dynamics evaluation and six observer states. SHAPE thus offers diagnostic capability without those observer requirements, with a remaining clean-performance gap to GMO.

Favorable regions also depend on representation and measurement quality. CWT's clearest advantage over Temporal occurs in intermediate Friction, while Temporal remains competitive in LoE and Vibration. Position-only perturbations leave SHAPE's predictions unchanged through its final measurement mapping, whereas the observer formulations degrade; velocity sensitivity limits this advantage. The empirical profiles therefore combine fault type, severity, operating stratum, representation, and measurement exposure. They distinguish high identification from relative gains in regions still affected by confusion or missed detection. Because fault type and severity are known when constructing these profiles, the regions characterize the evaluated cases rather than provide a selection rule for an undiagnosed fault.

\section{Conclusion}
\label{sec:conclusion}

SHAPE provides a construction for energy-based fault diagnosis without a dedicated dynamic observer in fully actuated, differentially flat pH mechanical systems with collocated actuation. It combines planned nominal port power with commanded effort and measured flow to generate physically interpretable residuals. With nominal references available, residual evaluation requires neither recursive observer updates nor measured-state dynamics calculations. The simulated manipulator results show that these signals support discrimination among the evaluated fault types through the CWT and shared-encoder CNN realization.

The comparisons characterize the diagnostic capability of this construction. Under common CWT processing, SHAPE achieves higher mean accuracy than the torque residual, while GMO provides broader clean diagnostic coverage. Severity--excitation profiles reveal fault-type confusion and missed detection, particularly for weak friction. CWT achieves higher mean accuracy than the temporal and Fourier SHAPE pipelines, while offline velocity perturbations cause pronounced degradation. The usefulness of the power residuals therefore depends on operating conditions, representation, and measurement quality.

Validation remains limited to simulated single faults at a specified joint. Further work should assess physical experiments, measurement errors propagated through the controller and plant, and complete-pipeline execution cost. Unknown fault locations and simultaneous faults also require evaluation. A longer-term direction is to connect the diagnostic signals to energy-aware control, including energy freezing; this requires online detection and analysis of detection delay and closed-loop safety beyond the present complete-trajectory classification.

\section*{Data Availability}
\phantomsection
\label{sec:data_availability}

The simulation and diagnostic code, frozen result tables, complete
regional maps, and configuration and provenance records are available at
\url{https://github.com/johnnyzxc5/SHAPE-paper-reproducibility}.
The repository documentation specifies the external artifacts and
environment requirements for classifier reevaluation and simulation
regeneration.

\section*{Acknowledgment}
OpenAI Codex~\cite{openaiCodex} was used to assist with language and clarity, preparing LaTeX code, and developing the project's GitHub reproducibility repository. The authors retain full responsibility for the manuscript, code, and reported results.

\bibliographystyle{IEEEtran}
\bibliography{references2} 

\end{document}